# A Novice Guide towards Human Motion Analysis and Understanding

Dr. Ahmed Nabil Mohamed
dr.ahmed.mohamed@ieee.org

**Abstract**

Human motion analysis and understanding has been, and is still, the focus of attention of many disciplines which is considered an obvious indicator of the wide and massive importance of the subject. The purpose of this article is to shed some light on this very important subject, so it can be a good insight for a novice computer vision researcher in this field by providing him/her with a wealth of knowledge about the subject covering many directions. There are two main contributions of this article. The first one investigates various aspects of some disciplines (e.g., arts, philosophy, psychology, and neuroscience) that are interested in the subject and review some of their contributions stressing on those that can be useful for computer vision researchers. Moreover, many examples are illustrated to indicate the benefits of integrating concepts and results among different disciplines. The second contribution is concerned with the subject from the computer vision aspect where we discuss the following issues. First, we explore many demanding and promising applications to reveal the wide and massive importance of the field. Second, we list various types of sensors that may be used for acquiring various data. Third, we review different taxonomies used for classifying motions. Fourth, we review various processes involved in motion analysis. Fifth, we exhibit how different surveys are structured. Sixth, we examine many of the most cited and recent reviews in the field that have been published during the past two decades to reveal various approaches used for implementing different stages of the problem and refer to various algorithms and their suitability for different situations. Moreover, we provide a long list of public datasets indicating both the importance of having those datasets available to the researchers and the importance of including their ground truth data. In addition, we also discuss briefly some important examples of these datasets. Finally, we provide a general discussion of the subject from the aspect of computer vision.

## 1- Introduction

Human motion understanding has been a desirable target for many researchers from different disciplines. Each discipline deals with a specific aspect of the problem. With such different motivations for researchers to bear in mind, many contributions have been made in different disciplines. Integrating those contributions together provide us with a better understanding of human motion. We will investigate various aspects of some disciplines (e.g., arts, philosophy, psychology and neuroscience) that are interested in the subject and review some of their contributions stressing on those that can be useful for computer vision researchers. Moreover, many examples are illustrated to show how the integration of concepts and results among different disciplines has been exploited by many researchers.

However, the main concern of this article is with the aspect of computer vision dealing with human motion analysis, recognition, and understanding. Therefore, this article tries to reveal the wide demanding and promising applications of the field. Then, we list various types of sensors that may be used for data acquisition. Next, we review various processes involved in motion analysis. Then, we review different taxonomies used for classifying motions. Subsequently, we review various processes involved in motion analysis. Afterwards, we explore different structures of previous surveys. Then, we spot many of the most cited and recent reviews in this area that are published in the past two decades to reveal various approaches used for implementing different stages of the problem and to review various algorithms and their suitability for different situations. Moreover, we provide a long list of so many public datasets indicating both the importance of having these datasets available to the researchers and the importance of including their ground truth data. Further, we also discuss briefly some important examples from semi-realistic action recognition datasets "simple datasets", more realistic and challenging action recognition datasets, interaction datasets and multi-view datasets. Finally, we provide a general discussion of the subject mainly from the aspect of computer vision and conclude with the overall trends.

The organization of the article will be as follows: In section 2, some disciplines and their contributions to the field of human motion understanding are discussed focusing on the contributions that can benefit computer vision researchers. Section 3 discusses human motion analysis, recognition, and understating in computer vision. Section 4 discusses several datasets, their importance and their use. Section 5 provides a general discussion of the field from the aspect of computer vision. Section 6 concludes the article.

## 2- Human Motion Understanding as an Interdisciplinary Field

Various motivations have pushed several generations of researchers from different disciplines towards studying of human motion from different aspects. There is no doubt that those studies on human motion were and are interdisciplinary. In this section, some of the disciplines that have contributed to the understanding of human motion will be discussed showing their contributions and mutual benefits between them.

**2-1 Visual Arts**

In prehistoric era, cave paintings were primarily about depicting game animals whether in profile or in motion. The earliest painters were very observing in depicting animals with great details, they chose the profile view that is completely informative about the animal shape in order to create convincing and realistic pictographic representations of them [153]. A very interesting study [100] revealed to us that cavemen were better at depicting quadruped walking than modern artists. The characteristics of quadruped walking are well known to the experts of animal locomotion since the pioneering work of Eadweard Muybridge in the 1880s. The study shows, after analyzing 1000 prehistoric and modern artistic quadruped walking depictions, that the error rate of quadruped walking illustrations of modern pre-Muybridgean period in respect of the limb attitudes presented, assuming that the other aspects of depictions used to determine the animals gait are illustrated correctly, was 83.5%, and it decreased to 57.9% after 1887, that is in the post-Muybridgean period. But, astonishingly surprising, the prehistoric quadruped walking depictions had the lowest error rate of 46.2% confirming that cavemen were more keenly aware of the slower motion of their prey animals and illustrated quadruped walking more precisely than later artists. Although, the rare human figures in prehistoric depictions appear abstract in comparison to overly emphasized details of animals, there are indications of existence of ritual cults, where adorants perform an act of adoration or invocation to a higher being [101, 153]. This act is always accompanied by a body posture or limbs motion to express or emphasize a desire or request to a higher being. These attitudes and gestures of adorants vary from standing, kneeling, sitting or squatting to prostrating, from inclining the head or bending the upper part of the body to raising one or both arms or extending them with the hands opened (above to the heavenly powers, down to the chthonic ones) or with spread fingers to uplifting the arms to or above the head, from clasping the hands, joining them to crossing above the breast, to touching an altar or idol to kissing them [101].

Ancient civilizations, such as those of Mesopotamia, Egypt, Greece, Roma, China, India, etc, revealed to us a good level of understanding of human and animal motion, and body poses. Many sculptures, reliefs, paintings, drawings and other artifacts show us different activities, religion practices, and events existed in their time. For example, ancient Egyptian art was mainly concerned with depicting religious beliefs and practices, magic practices, and social life; It used detailed depictions of gods, human beings and battles (e.g., temple walls depicted the king making offerings to the gods, and riding in a chariot over wounded, dying, and defeated foes) to reveal the Egyptians beliefs about the world and their attempts to understand it [107, 108]. Motion in early Chinese art was also depicted as an occasional escape from the spells of the pure ornamental style, e.g., all good Han representation is vivid and alive and draws its material from the wide interest in human and animal activity [109]. Hellenistic art is another example to mention, it was very interesting in expressing human body in great details, its poses, and its movements. This art paid also a great attention to facial expressions such as suffering, anxiety, etc.

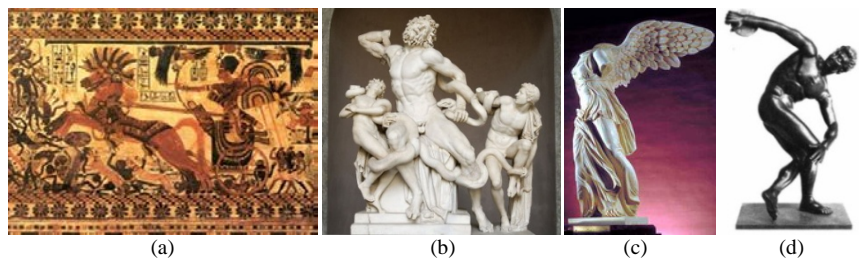

(a) (b) (c) (d)

Fig. 1 Motion understanding in arts of ancient civilization: a) King Tut riding his chariot and carrying his bow in a battle chasing Nubians, b) sculpture of Laocoon and his sons suffering and trying to loosen from snakes, c) Winged Victory statue giving a sense of motion due to the dramatic effect of her sweeping draperies, d) discus thrower.

Many famous artists portrayed human motion in a convincing realistic way. For example, Leonardo da Vinci's (1452 - 1519) lust for human motion understanding appeared very clearly when he stated in his sketch books "it is indispensable for a painter, to become totally familiar with the anatomy of nerves, bones, muscles, and sinews, such that he understands for their various motions and stresses, which sinews or which muscle causes a particular motion" [102]. Da Vinci showed in his sketch books, beside very detailed models of human anatomy, impressive level of detail in modeling human shape and motion as he also contained detailed studies about kinematic trees (known today as kinematic chains) of human motion [102]. Michelangelo (1475 - 1564) expressed his connection to showing motion in his artworks by stating "How wrong are those simpletons, of whom the world is full, who look more at... color than at the figures which show spirit and movement." [156]. Edgar Degas (1834 - 1917), a French artist who is especially identified by depicting dancers in more than the half of his works, stated "People call me the painter of dancers, but I really wish to capture movement itself." [157]. Ernst Ludwig Kirchner (1880 - 1938), a famous German Expressionist Painter, was very fond of depicting human figures with a sense of motion. He wrote, "Why didn't those worthy gentlemen [other painters] paint real life? Because it moves, that's why. They neither see it nor understand it. And then I thought -- why shouldn't I try? And so I did." "I drew in the streets and squares, in taverns, in the circus, in cafes. Anywhere I could see people in motion". He also said about Rembrandt van Rijn, an earlier famous Dutch painter,

"First of all I needed to invent a technique of grasping everything while it was in motion, and it was Rembrandt's drawings...that showed me how" [154].

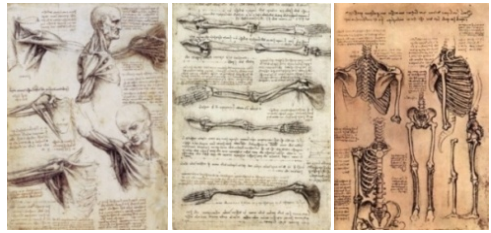

Fig. 2 Da Vinci's studies of neck, shoulder, arm, and skeleton [155]

Étienne-Jules Marey (1830 - 1904), a French scientist and chronophotographer who was interested in animals and human locomotion, published a book about motion studies where data had been collected by various instruments, and used his photographs for the purpose of illustration and analysis. He invented special cameras allowing for recording several phases of motion in one photograph [102], see the top left of fig. 3 (the seer of this photo recalls, at once, motion energy image "**MEI**" and motion history image "**MHI**" developed by Bobick and Davis [52]). He also captured movements of humans, by a single camera on a single plate, while wearing black suits with white strips in front of black panels, see the top right of fig. 3. A similar technique is used by video-game designers to translate the natural movements of human actors into virtual computer graphics characters [103]. Eadweard J. Muybridge (1830 - 1904), an English American photographer, who was interested in studying motion based on multiple images, set up a series of cameras for recording fast motion of a horse when galloping and trotting in order to settle a debate about whether a horse may have all of its hooves off the ground simultaneously or not. He also studied motion of other animals and humans, and published his book, animal locomotion, which contain over than 100,000 photographs in 1887 [103]. The work of Muybridge and Marey inspired many artists such as the futurists, but the Italian Futurist art movement, which came as a response to the rapidly developing technologies, aimed to depict continuous movements rather than the images of sequential sharp phases of motion that Muybridge and Marey had mastered [103, 154]. Muybridge and Marey pioneered cinematography where capturing of motion in movies became very accessible [102]. In present, some consider the use of 3D motion capture technology to obtain the trajectories of human movements is a way to show the beauty of the nature of our movement [111].

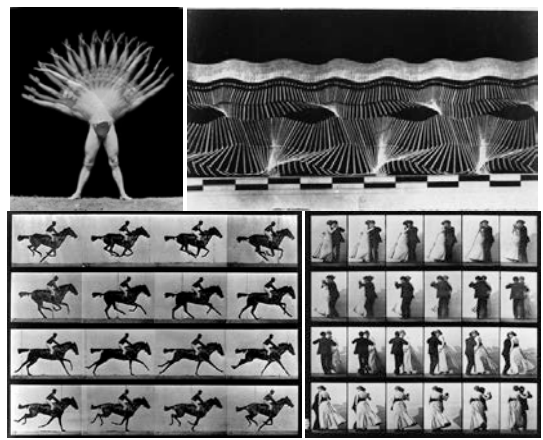

Fig. 3 The top photos are produced by Marey and the bottom ones by Muybridge

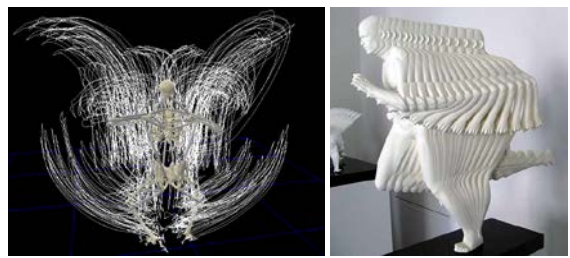

Fig.4 Human motion in modern arts: a) the signature of Indian dance captured in 3D [110], b) a sculpture captures sequences of human movements in space and time made by Peter Jansen (a contemporary artist)

It is obvious that the simplest way to represent motion in a painting or a photograph is to choose the right moment to paint or shoot, but of course, painters have the advantage of incorporating many different moments in one image but they need keen eyes to observe well [103]. Cutting [104] discussed how artists, photographers and scientists represent motion in a static image or a piece of artwork. He categorized their representations of motions into five different types, which are: 1- dynamic balance and broken symmetry (e.g., see fig. 1d), 2- multiple stroboscopic images (e.g., see the top photos of fig. 3), 3- plasticity, affine shear and forward lean (i.e., a moving object is depicted as leaning into its direction of movement; that works for even inanimate objects such as a car or a train), 4- motion blur (e.g., taking a photo while the object is moving), and 5- action lines, zip ribbons and vectors (i.e., the depiction of motion direction is appeared through an array of lines or a zip ribbon attached to the object). Motion can also be represented in a series of static images to indicate events in longer time periods (e.g., some ancient Chinese handscrolls, motion-picture-like quality of the gāmi at-tawārīH [110], Donald Duck comics, etc.). Computer visualization researchers may use NPR (Non-Photorealistic Rendering) techniques to convey change in the volumetric data over time. For example, Alark et al. [105] proposed new techniques inspired by the fifth category of Cutting's categorization, mentioned earlier, to convey change over time more effectively in a time-varying dataset. He used techniques based on speed lines, flow ribbons, opacity modulation and strobe silhouettes. He also claimed that his work is valuable to domain experts, who would prefer to watch temporal summaries of time-varying data instead of watching hours of animations/surveillance videos which may have no interesting activities. Cavanagh [106] deemed that artists may be of great help in understanding visual perception and recognizing everyday activities and scenes.

In this subsection, a brief review has been presented about how motion, in general, and human motion, in particular, has grasped the attention of mankind along his history through various depictions whether before recorded history when he was interested in hunting animals or practicing ritual cults, or in ancient civilizations where depictions of religious beliefs, battles, kings and rulers in idealized features and poses, social life and different activities of human and animals took place. It also has been shown how human motion grasped the attention of many famous artists and photographers, and how they developed techniques and invented devises for representing motion in their artworks, trying to analyze and understand it or unveiling hidden beauties of our motion. In the end, a brief explanation has been introduced about how motion is represented in visual arts and sciences, and how theses representations can be of great help for different researchers such as computer visualization researchers, psychologists, neuroscientists, etc.

**2-2 Philosophy**

Many Philosophers have questioned about human and animal motion, amongst them was Aristotle (384 BC – 322 BC), the ancient Greek philosopher, who wrote a book [112] on the gait of animals where he discussed parts that are useful to animals and humans for movement in place (locomotion). He discussed things like why each part is as it is, and what differences between parts both in one and the same creature, and made comparison of the parts between creatures of different species. He investigated what fewest points of motion are necessary for animal and human progression, why sanguineous animals have four terminals not more but bloodless animals have more than four, why, in general, some animals are footless, others bipeds, others quadrupeds, and why all have an even number of feet, if they have feet at all, why progression depends on even number of points, why do quadrupeds move their legs crisscross, and why some animals move with the whole body at once, e.g., jumping animals, and others move one part first and then the other, e.g., walking animals. He also discussed why man and birds are bipeds, why they have an opposite curvature of the legs, why man bends his arms and legs in opposite directions when moving, and why man is the only erect walking animal while young children cannot walk erectly. Aristotle, in his quest for answers to these inquiries and many other inquiries that are not listed here, said that we must assume that the originals of movements in place are thrusts and pulls, and it is only accidentally that what is carried by another is moved. He also said that the boundaries by which living beings are naturally determined must be taken into consideration; these boundaries are superior and inferior, before and after, right and left. As Aristotle was not just a philosopher, he was also a polymath, he used, for example, his knowledge of Pythagorean theory to explain why the leg at rest must bend at its knee when man is moving, and thus, he integrated his knowledge in mathematics to answer some of his philosophical questions. He also stated in one of his experiments that "if a man were to walk parallel to a wall in sunshine, the line described by the shadow of his head would be not straight but zigzag becoming lower as he bends, and higher when he stands and lifts himself up", this saying was reinforced by Johansson in one of his MLD experiments [121] when he displayed typical motion paths of seven parts (shoulder, elbow, hip, wrist, knee, left and right ankles) representing a walking person, see fig. 5.

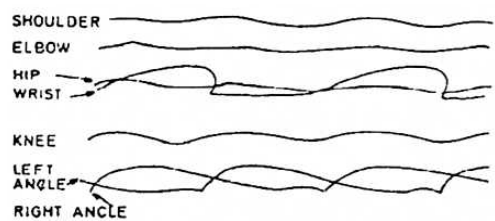

Fig. 5 Typical motion paths of seven parts of a human body when walking [121]

In another direction, several philosophers, especially phenomenologists, claim that we understand actions of others through an internal act that recaptures the sense of their acting [126]. Recently, collaboration between neuroscientists and philosophers has led to integrate phenomenological descriptions with physiological descriptions of human action.

On the other hand, philosophy has also benefited other disciplines towards recognition and understanding of human activities and behaviors. For example, in the 1980s, early artificial intelligence researchers had adopted the term "ontology" from the field of philosophy to enable the modeling of knowledge about some domain, real or imagined. They recognized the applicability of the term from the mathematical logic and argued that it could be used to build computational models that enable certain kinds of automated reasoning. In fact, some researchers consider computational ontology as a kind of applied philosophy [95]. Let us, first, take a fast look at the meaning of ontology in philosophy, and then explain its meaning in the context of computer and information sciences, and finally discuss its use in computer vision for recognizing human activities. Ontology as a branch of philosophy is defined as the science of "what is" that concerns with the study of the nature and relations of being, or the kinds of things that have existence, or reality. Traditionally, it is considered as a synonym or a major branch of "metaphysics", a term used by early Aristotle's students to refer to what Aristotle himself called "first philosophy". Sometimes, the definition of ontology is extended to refer to the study of what might exist [97]. In computer and information sciences, an ontology refers to a set of representational primitives that are used to model a domain of knowledge. These representational primitives are typically classes, attributes and relationships [95]. Gruber [96] defined ontology as an explicit specification of a conceptualization and presented a preliminary set of criteria to design and evaluate ontologies for the purpose of supporting knowledge sharing and interoperability. The five criteria principles are: clarity, coherence, extendibility, minimal encoding bias, and minimal ontological commitment. In computer vision, ontologies serve as centralized representations of activity definitions for modeling and recognizing human activities. Thus, they standardize activity definitions, facilitate portability to specific scenarios, enable interoperability of different systems, and allow reusing and comparison of system performance. In a word, we can say that an ontology of a human activity describes objects, environment, interactions between them, and the events sequence. Several researchers have designed ontologies for specific scenarios in visual surveillance, and as a result, the video Event Challenge Workshop was held in 2003 to unite these efforts and to build a common knowledge base of domain ontologies. As a consequence of this workshop, six domain ontologies were defined for visual surveillance: 1-perimeter and internal security, 2-bank monitoring, 3-store monitoring, 4-metro monitoring, 5-railroad crossing surveillance, and 6- airport-tarmac security. The workshop led also to the development of two formal languages: VERL (Video Event Representation Language), and VEML (Video Event Markup Language). The first one provides representations of complex events in terms of simpler subevents, and the second one is used to annotate VERL events in videos [94].

In conclusion, we can say that philosophy has contributed to the field of human motion analysis and understanding, and that there are mutual collaborations between philosophy and other disciplines to further the advances in the field.

**2-3 Psychology**

It may be interesting to know how we develop our perceptual ability to acquire information or knowledge about different kinds of actions or objects. In the past, there was a debate among psychologists about whether this ability is innate or whether it emerges after birth as a result of specific experiences. This debate was known as nature-nurture controversy. But there is an important factor that is not directly addressed by either of these two views which is biological maturation, since for example, our vision system lacks maturity at birth. Therefore, there are three key points that should be considered carefully to understand the perceptual development, which are: physical maturation, the role of experience and learning, and the innate sensitivity to acquire information [113].

Gestalt principles of perceptual organization [118, 119] that account for the visual perception of spatial configuration may help to tackle many problems faced by computer vision researchers. Some classical and new principles of perceptual grouping [115, 116], a very important kind of organizational phenomenon are mentioned here briefly as follows:

- Proximity (nearby elements are grouped together),
- Similarity (elements that look the same are grouped together, e.g., same color or size or orientation),
- Common Fate (elements that move in the same way are grouped together),
- Good Continuity (straight or smoothly changed contours perceived as the same object),
- Closure (an incomplete object is perceived as a whole),
- Symmetry (interpretation of objects as symmetrical or simple forms),
- Generalized Common Fate (elements that become darker or brighter simultaneously (same luminance change), even if they have different luminance, are grouped together),
- Synchrony (elements that change simultaneously are grouped together, whatever the changes are in the same direction or not), synchrony is considered as an even more general form of common fate,
- Common Region (elements that lie within the same bounded area are grouped together, e.g., spots on a leopard skin are perceived as elements on the surface of a single object "the leopard"),
- Element Connectedness (distinct elements that share a common border are grouped together, e.g., bristles, metal band, and handle, all constitute a single object known as a "paint brush"),

- Uniform Connectedness (an image is initially partitioned by the visual system into a set of mutually exclusive connected regions having uniform or smoothly changing properties, such as luminance, color, texture, motion, and depth).

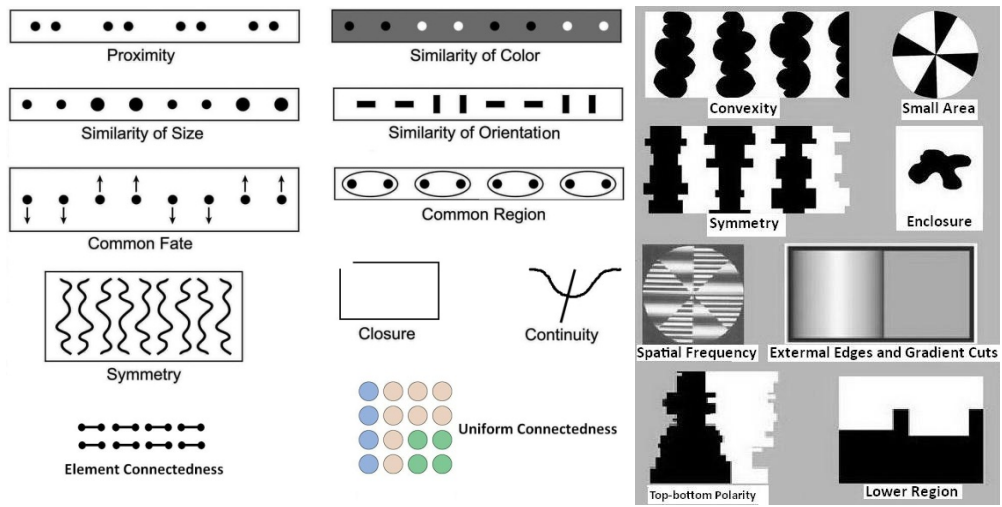

Fig. 6 Illustrations of some gestalt grouping and figure-ground principles (adapted mainly from [115, 117]).

Another important kind of perceptual organization is figure–ground organization [115, 117]. Regions of the image that are characterized by convexity, symmetry, small area, or enclosure (being surrounded) are more likely to be perceived as figures than contiguous regions that are concave, asymmetric, larger in area, or surrounding. These properties are called classic image-based configure principles of figure-ground organization; others may call them classical geometric configural properties. New image-based configure principles of figure-ground organization include:
- Lower Region (The lower of two regions divided by a horizontal border is more likely to be perceived as the figure than the upper region),
- Top-Bottom Polarity (regions that are wider at the bottom and narrower at the top are more likely to be perceived as figures than regions that are wider at the top and narrower at the bottom),
- Extermal Edges and Gradient Cuts (when shading and texture gradients are used to depict a self-occluding edge "an extremal edge" along one side of a border but not the other, the side with an extremal edge gradient is almost invariably perceived as being closer to the observer than the opposite side of the edge),
- Edge Region Grouping (if figure-ground organization is determined by an edge grouped to the region on one side more strongly than that on the other, then any grouping factor capable of relating an edge to a region is also considered as a figure-ground factor),
- Spatial Frequency (a region filled with a high spatial frequency pattern is more likely to be perceived as the figure than a contiguous region filled with a low spatial frequency pattern).

Other non-image-based influences on figure-ground perception include: 1- Past Experience, 2- Attention, 3- Direct Reports (regions that portrayed portions of familiar objects were more likely to be perceived as figures when they were upright than when they were inverted), 4- Indirect measures (such as response time). Shape description is closely tied to figure-ground organization because it is common to perceive the edge between two contiguous regions as a boundary for only one of them, which is the figure that appears to have a definite shape while the ground appears shapeless near the edge it shares with the shaped figure (for a discussion of various models that are used to represent and recognize shapes such as templates, feature, Fourier, structural description, etc, refer to Pinker [124]). Of course, figure-ground organization is also related to depth cues due to the fact of the figure being closer to the observer than the ground. In fact, some may consider figure-ground principles as a subset of depth cues. With the importance of perceptual organization principles in mind, many computer vision researchers try to engage these principles in their problem-solving. For example, Martin et al. [98] presented a database of natural images, which was segmented by human operators, to be used by computer vision researchers in evaluating segmentation algorithms and then calculated probability distributions associated with the gestalt principles of proximity, similarity of intensity, and convexity of regions in order to indicate that the segmentation process is based on ecological statistics. Shah [46] believes that employing dynamical perceptual organization would be helpful in understanding human behavior from motion imagery. This would be accomplished by extending the work on perceptual organization for single images to video sequences in order to discover relations between 2D motion and 3D motion.

It may also be of importance for computer vision researchers to know how we perceive depth biologically in order to help them in solving different problems. Humans can perceive depth through three different types of information which are kinematic information, binocular information, and static-monocular information [113]. Kinematic information is acquired whenever the perceiver or the object(s) are in motion. Examples for Kinematic information cues include:
- Accretion and deletion of texture (it happens when one object covers another one),
- Expansion or contraction of texture elements, optical expansion, (it happens when an object moves away from or close to the perceiver),
- Motion parallax (it happens when nearby objects appear moving more quickly than distant objects).

Binocular information can be acquired through moving our two eyes to place the image of an object on our fovea, the region with best acuity, or after binocular disparity calculation that uses the available information at the retinal level. Static monocular information is available under static conditions and do not require both eyes of the perceiver, this class of information is also called pictorial cues because many of them were described by Leonardo da Vinci for portraying depth in paintings. These cues include:
- Texture Gradients (assuming that texture on the same surface are equal in size, then parts of the surface with larger texture elements are closer to the perceiver than those parts with smaller texture element),
- Linear Perspective (e.g., parallel sides of a road appear to converge in the distance),
- Shading,
- Interposition (also called "overlap", it may happen when an object occlude part of another object),
- Relative Size (It may happen when objects of the same size decrease in size with distance),
- Familiar Size (objects with familiar sizes to the perceiver provide a cue about their distances from him).

Gibson theory about invariance in visual perception, which states invariants of the energy flux at the receptors of an organism are the basis of the organism's perception of the environment and correspond to the permanent properties of the environment, have pushed many psychologists to seek help from mathematics discipline about using the term invariance, which is rooted in mathematics, in perception. Cutting [120] discussed four assumptions that underlie the application of the term to perception and assessed their validity. The first assumption is that mathematics is an appropriate language for describing perception, this assumption is based on two ideas: the world is best described in some form of mathematics, and that the human mind is attuned to that description since it is attuned to the spatial layout of the world. The second assumption is that mathematical truths are transportable into perception without change in meaning, and hence assessing whether the term invariance means the same thing in mathematics as it has come to mean in perception must be validated carefully and that acquires more background about the term usage in mathematics. The third assumption is that mathematical imports are useful in explaining perception, for that to be happen, overgeneralization should be avoided. The fourth assumption is perceptual invariants are absolute as in mathematics, Cutting believed that this assumption to be false.

Perception of human actions depends upon multiple sources of information including sensory, motoric and affective processes [114]. Psychological studies indicate, without any doubt, the significant role of low-level motion processing (bottom-up approach) in perceiving human motion but there are also other studies that show the influence of high-level processing (top-down approach) in supporting the perception of human motion. In addition, several psychological studies indicate the important roles of both form and motion in the perception of human action. Inspired by these psychological studies, Chen and Fan [99] designed a computational framework to guide human motion segmentation by combining the visual perception processes of bottom-up (feed forward) and top-down (feedback). They claimed that their model can deal with cluttered scenes and can also decide when to stop the cascaded merging processes in the bottom-up layers. Concerning with motoric processes, it is important to note that our keen ability to perceive actions of other people is, in part, due to the massive experience that we have accumulated over our lifetime in planning and executing different kinds of actions, and thus, sensory representations used during action perception overlaps with motor representations used in planning and executing actions. Several psychophysical studies support the above idea by confirming the following: 1- action perception and action production share common representations so that an observer's own activities influences the observer's perception of the activities of other people, 2- Observers demonstrate maximum sensitivity to actions most familiar to them and reduced sensitivity to the actions unfamiliar to them, 3- Patients with congenital and disease-related disorders have their perceptive mechanisms and/or motor behavior affected. In summary, one's own actions affect his/her perception of actions produced by others. Many other psychophysical studies have also shown that a wide range of socially relevant characteristics can be deduced from highly degraded depictions of human actions such as a performer's identity, dancing ability, angry walker, openness, social dominance, vulnerability to attack, and ability to deceive, and thus, an expectation about the influence of social processes in action perception is a reasonable one.

Psychologists used many techniques for studying the robustness of perception of human motion. One of these techniques was devised by Johansson [121] where he represented the motion of a human body by a few bright spots describing the motion of the main joints; this technique is known as Moving Light Displays (**MLD**) or point light animation of biological motion (simply PL animation), see fig. 7. He found that 10-12 of these elements, in adequate motion combinations, are sufficient to evoke us with the impression of human walking, running, etc. By using the same technique, subsequent studies have shown that even the sex and the identity of the walking person can be identified [122, 123]. Many variations have been applied to this technique [114], such as, using it to recognize facial expressions, applying noise dots that are more in number than the dots marking the person, blurring the dots, etc, and it has been found that the perception of human motion is still remarkably robust. It has also been found that points marking the wrists and ankles are of great importance when recognizing the walking direction and those points marking elbows, knees, shoulder and hips are significant for the perception of human motion when noise exists. It has been shown also that points marking the joints are more compelling for the perception of human motion than those placed on the limbs. But perception of human motion failed when inverting the PL animations or when they are viewed under dim light conditions. Other techniques are also successfully used for testing the perception of human motion such as connecting the points in MLD with visible line segments to create stick figures or by using motion capture systems, key frames, etc.

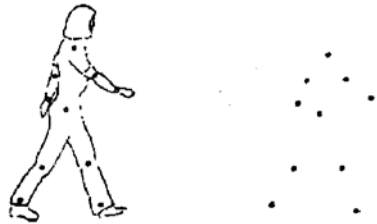

Fig. 7 A contour of a walking person and corresponding bright dots configuration [121]

In this subsection, we have introduced a brief hint about perceptual development in humans to acquire the external world. Then, we have discussed a number of principles described by gestalt psychologists to account for the perception of spatial configuration. Next, different kinds of information through which humans can perceive depth were explained. Then, a theory by Gibson about invariance in human perception was introduced, indicating how psychologists may use the term invariance from its mathematical origin and apply it in perception. Moreover, we have presented brief hints about sensory, motoric and affective processes involved in human perception. Finally, we have briefly shown some techniques that are used to test the perception ability robustness of human motion. (Note: throughout this subsection, we have provided some examples to indicate the integration of results between psychology and computer vision).

**2-4 Neuroscience**

Several neurophysiological studies have found that there are some neurons in the superior temporal sulcus (**STS**) that respond selectively to human activities, or in other words, to human forms and motions, these findings have been confirmed by other experiments in brain imaging such as positron emission tomography (**PET**) and functional magnetic resonance imaging (**fMRI**) [114, 127]. The STS is able to integrate form and motion information because it represents a point of convergence for the ventral (what) and dorsal (where) visual streams that are responsible for processing form and motion information, respectively. Many other studies indicate that the STS plays a fundamental role in the perceptual analysis of social and emotional cues and it may even be used to determine the social significance of an action [114].

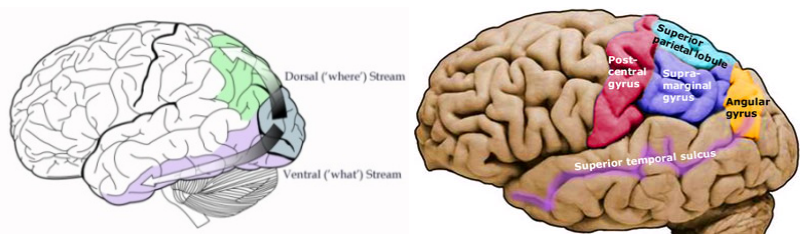

Fig. 8 The left: dorsal and ventral visual streams of a human brain, the right: STS in a human brain

Other remarkable neurophysiological studies discovered another category of visually activated neurons that are called "mirror neurons" which fires when an animal acts and when it observes another individual executing the same act, these neurons seem to generalize over varying executions of the same action. Mirror neurons were first discovered in macaque monkeys in an area of the brain called F5, and have later been also found in the inferior parietal lobule [126, 128, 114]. Neurophysiological (Transcranial Magnetic Stimulation "**TMS**", Magnetoencephalogram "**MEG**", and Electroencephalogram "**EEG**"), and brain-imaging (**PET** and **fMRI**) experiments provided strong evidence that a mirror neuron system exists also in humans [128, 130]. Many experiments show that the properties of the mirror neuron system in humans indicate the existence of a mechanism that may underlie a variety of functions such as action understanding and imitation [126, 128], intention understanding and empathy (vicarious experiencing of the feelings, thoughts, or attitudes of another) [129].

Several neurophysiological models have been developed for the recognition of objects and/or their movements. For example, Giese and Poggio [127] developed a feedforward hierarchical neurophysiological model, where form and motion information are analyzed in parallel, to account for many experimental findings in biological movement recognition. The model uses learned prototypical patterns that are stored in specific neurons in the ventral and dorsal pathways of the visual system for the learning of biological movements. Unfortunately, the model lacks, among other things as declared by the authors, top-down attentional influences on the recognition of biological movement especially in visual clutter and noisy scenes.

Marr [138, 139], one of the most influencing neuroscientists, integrated his insight and knowledge from mathematics, neurophysiology, and psychology into computer vision and image understanding. He emphasized in his work that a general theory in vision and behavior sciences would not be complete if one turns a blind eye to the mathematical rigor for the sake of specific findings. Thus, he introduced the concept of computational theory in the visual perception (It is worth noting here, that he considered the invariance perception theory of Gibson, discussed in the previous subsection, as the nearest attempt to what he called the computational theory). Moreover, he stressed on that seeking help from computational vision to explain some phenomena in human vision would be rather helpful, and vice versa. In his quest for understanding perceptual information processing, he proposed in his book [135] that an information system must include three different levels. The top level concerns with the computational theory involved, which describes what is done (the mapping from the input to the output) and why (the constraints of this mapping that solves the problem). The middle level concerns with the appropriate input and output representations and the algorithm used for transforming the input to the output. In other words, the middle level describes how the mapping is done. The last level concerns with the physical implementation of the system. Another interesting idea, which Marr described also in his book [135], is a representational framework for deriving shape information in visual processing. His three stages of vision for object recognition are: 1- primal sketch of the visual scene (which detects image intensity changes and identify primitives such as blobs, edges, boundaries, etc), 2- 2.5D sketch (is intended to groups pixels into a coherent representation of an object's boundary and represent the orientation and rough depth of the visible surfaces as well as discontinuities, it uses shape-from-x methods to recover the lost dimension, where x stands for visual cues, such as stereo, motion, shading, texture, etc), and 3- 3D model which describes shapes and their spatial organization in an object centered coordinate. Marr constructed, with others, representational models of static and moving 3D objects, which have natural or canonical axes based on a shape's elongation (e.g., humans), for the purpose of recognition [136, 137]. He also introduced with Hildreth a laplacian of gaussian edge detector that performs pretty well compared to other detectors like Sobel, Prewitt, etc [45]. Although Marr's theory about object recognition is one of many other things that have not survived but his influential work has shaped a new way for studying human and computer vision, making a lasting impact on vision research. In recognition to his inspirational work, one of the most prestigious honors for a computer vision researcher is named after him, the Marr Prize, which is given by the committee of the IEEE International Conference on Computer Vision (ICCV) that is ever considered with the IEEE Conference on Computer Vision and Pattern Recognition (CVPR) as the top conferences in computer vision.

Another example of how usefulness is the integration of results among different disciplines (here, we mean mathematics, neurophysiology, and psychophysiology disciplines) that can benefit towards the understanding of human motion is shown in Marcelja [131] where he argued that the representation of an image in the visual cortex corresponds closely to the Gabor theory [132] of analyzing signals. He pointed out that the visual cortex representation of an image must involve both spatial and spatial frequency variables due to the measured receptive field profiles of the simple cortical cells and their spatial frequency tuning characteristics. He also showed that the curves of these two properties (measured receptive fields and measured spatial frequency tuning characteristics) conform in a consistent manner with the Gabor representation where an arbitrary function is expanded in terms of symmetrical and antisymmetrical elementary signals. He said that although many mathematical explanations may provide descriptions of the cortical cells responses in both spatial and spatial frequency domains but the Gabor scheme has the advantage of dealing with the subsequent problem of pattern recognition because of its optimal localization properties in both spatial and frequency domain. Daugman [133] generalized the application of Gabor's one-dimensional scheme to 2D spatial vision by introducing an optimal 2D filter family that is later referred as 2D Gabor filters by Jones and Palmer [134] where an evaluation of the 2D Gabor filters model of the visual cortex is well established. Gabor filters have been used in many applications such as texture representation and segmentation, feature extraction, edge detection, target detection, stereo disparity estimation, retina identification, image representation, etc.

In this subsection, we have referred to the significance of STS in perceiving human activities and the perceptual analysis of social and emotional cues. Next, we have pointed to the importance of mirror neurons in understanding and imitating actions, intention understanding and empathy. Then, a neurophysiological model for recognizing human motion was mentioned. In the end, two examples were given to illustrate the benefits of integrating results among different disciplines such as mathematics, neurophysiology, and psychophysiology.

**2-5 Summary**

Although, this article is, of course, not the first one that indicate the interdisciplinarity of the field of human motion understanding, but it approaches the point from a new perspective stressing on the following issues:
- Several disciplines have different interests in human motion analysis and understanding, we have briefly viewed few of them, but there are still many other disciplines with other different interests, such as, kinesiology or biomechanics (where modeling of the human body is the primary goal in order to study, explain, and improve its mechanical functions), computer graphics (where synthesis of human motion is essential for developing realistic models of human body in order to be used in applications such as video games, movies, etc), choreography (design and representation of human movements in dancing, ballet, gymnastics, underwater dancing, theater, show choir, fashion show, marching bands, martial arts show, beatings in movies, etc. Labanotation is one of the notation systems that is used for representing human movements in choreography [Badler and Smoliar [62] discuss notation systems and representations of the human body and its movement indicating how Labanotation could be used to animate a realistic human body on a graphics display]), etc.
- Integrating results among different disciplines leads to enriching these disciplines with helpful ideas and methods, provides a better understanding and leads to more advances in the field.
- Indicating some contributions that can be of importance to computer vision researchers.

### 3- Human Motion Analysis, Recognition and Understanding in Computer Vision

The ultimate goal of computer vision is to understand the scene correctly through various steps of acquiring, processing, analyzing and understanding different kinds of information obtained by different kinds of sensors. Human motion analysis and understanding is one of the very hottest topics within computer vision. We will focus here on vision-based human motion analysis.

In this section, we will review some of the very potential and demanding applications concerning human motion analysis and understating in computer vision. Then, we list various types of sensors that may be used for data acquisition. Next, we list various processes in different levels that are involved in human motion analysis. Then, we review different taxonomies used for classifying motions and for structuring different surveys. Finally, we spot many of the most cited and recent surveys in this area that are published in the past two decades to reveal various approaches used for implementing different stages of the problem and to review various algorithms and their suitability for different situations. The surveys are listed in a chronological order to reveal the progress of the field.

### 3-1 Applications

Human motion analysis and understanding has gained much interest and research in computer vision due to the wide range of demanding and promising applications. A brief review of some of these applications in different domains is listed as follows:
- Smart or Automated Surveillance: as the attention of a good, competent, and dedicated vigilant person decreases after 20 minutes into unacceptable levels due to boring and hypnotizing nature of monitoring video scenes [63], and with the growing numbers of cameras covering vast areas, the human operator becomes more costly and unreliable (e.g., a survey of CCTV (Closed Circuit Television) systems in one London borough revealed that over 75% of the institutions that apply the CCTV system had no dedicated monitoring staff [64]). Thus, the need for automated surveillance systems turns to be very urging. Some applications of smart surveillance are: suspicious behaviors [65] and unlikely events [66] detections; understanding and describing human behaviors in dynamic scenarios (e.g., monitoring activities over a complex area using a distributed network of active video sensors [67]); access control in special areas such as military bases and important governmental units where the system should automatically obtain biometric features of the visitor and then decide whether the visitor can be cleared for entry; person-specific identification at a distance which can help the police in chasing and catching suspects by placing surveillance cameras in locations where suspects may appear such as subway stations and casinos; safety monitoring: e.g., detecting drowning in swimming pools [68]; consumer demographics in shopping malls; crowd statistics [69, 70] and pedestrian congestion in public areas such as stores and travel sites, security applications in places such as banks [71], department stores, office buildings, parking lots, shopping centers, public transportation [72], borders, and homes.
- Behavioral Biometrics: recognizing humans based on their behavioral cues (e.g., human gait [73, 74], length, facial features, etc) does not require subject cooperation or intervention in their activities.
- Human-Computer Interaction: enables the user to control and command, e.g., gesture driven control, eye gaze tracking [75], speech recognition, sign language translation and understanding, signaling in high noise environments such as factories and airports, perceptual user interfaces [76] that allows a computer user to interact with the computer without having to use the normal keyboard and mouse by giving the computer the capability of interpreting the user's movements or voice commands.
- Virtual Reality: where the user is able to interact with a computer-simulated environment, e.g., training of military soldiers, firefighters and rescue squads by learning in simulated environments. Virtual reality has also many applications in game and entertainment industries.
- Smart Environments: where extracting and maintaining awareness of a wide range of events and human activities take place, e.g., monitoring interactions of participants in a meeting room [77].
- Games Industry: several games use the gesture-based interactive technology where motion capture is employed to enable interaction between a player and a game through non-intrusive body movements. For example, the famous Microsoft Kinect Xbox [78, 79].
- Entertainment Industry: precise motion-capturing is used in Sci-Fi movies to replace actors with animation

characters (digital avatars) [80].
- Video Annotation, Indexing, and Retrieval: as the number of videos increases rapidly due to the magnificent progress in capturing and recording technologies, accompanied by decreasing costs of cameras and storing media, the need to annotate and index various kinds of videos including personal videos, sports videos, news broadcasting, movies, surveillance videos, etc, becomes very insisting to save time and labor in retrieving them in a more easy, fast and convenient way, e.g., acquiring a certain highlight in a soccer game [81] or in news broadcasting. A review of recently developed information retrieval techniques can be found in [82].
- Physical Therapy: e.g., non-intrusive capturing of normal and pathological human movement [83], diagnosis of orthopedic patients, etc.
- Sports Motion Analysis: analyzing different sports such as the soccer sport [84] where verification of the following issues may be addressed such as referee decision, tactics analysis, automatic highlight identification, video annotation and browsing, content based video compression, automatic play summarization, advertisement insertion, player and team statistic evaluations, etc.
- Human motion analysis and synthesis: acquiring accurate movements of athletes, dancers, fighters, etc, for performance analysis, evaluation and enhancement and for training purposes.
- Robotics Learning by Imitation: e.g., robots may be used to provide daily social services such as setting up or cleaning dinner tables. Bandera et al. [85] provide a survey on vision-based architectures for robot learning by imitation.
- Assisted Living or Proactive Services: assisting disabled people, elderly people, children as well as normal people, e.g., fall detection systems [86] that monitor the person's movements and call the corresponding emergency center if it detects a falling person. Chaaraoui et al. [87] provide a review on human behavior analysis for ambient-assisted Living.
- Intelligent Driver Assistance Systems: where the assisting process must be very efficient and in real time, e.g., monitoring driver awareness [88], sleep detection, airbag system control, predicting driver turn intent [89], pedestrian detection [90], etc.
- Autonomous Mental Development [91]: this includes studying how the human brain develops its mental capabilities through examining autonomous real-time interactions with its environments using its own sensors and effectors, e.g., study the cognitive learning process of young children [92].
- Video Compression: e.g., using model-based coding allows very low bit-rate compression [93].

These applications vary in their requirements (e.g., human modeling, real time processing, video resolution, controlled or uncotrolled environmets, active or passive sensing, types and number of sensors, performance robustness and accuracy, etc) to achieve human motion analysis and recognition.

**3-2 Types of Sensors Used for Data Acquisition**

As the main goal of computer vision is to derive information from the observed scene, several types of sensors can be used for data acquisition such as: still cameras, video cameras, night-vision cameras, markers on the human body, special body suits and gloves, laser rangefinder (used to determine the distance to an object by applying a laser beam, it may operate on technologies such as time of flight), light detection and ranging "LiDAR" (is a remote sensing technology that measures distance by sending pulses of laser light that strike and reflect from the object surface, it could be used in robotics in order to percieve the surrounding environment), structured light (calculates the depth and surface information of the objects in the scene by projecting a known pattern of pixel, e.g., grids, and measure the deformation), sound navigation and ranging "Sonar" (which uses sound propagation to detect objects), radio-frequency identification "RFID" (used to transfer data, for the purposes of automatically identifying and tracking tags attached to objects), radiometers, millimeter wave radar, microwave radar, synthetic aperture radar, tomographic motion detection, x-ray sensors (can give us a complete image of a whole human body without any occlusions), inertial measurement units (electronic devices that measure object's velocity, orientation, and gravitational forces, using a combination of accelerometers and gyroscopes, sometimes also magnetometers), fiber optic sensors (used to measure strain which can, in turn, be used to recognize body postures), pressure-sensitive foam sensors (to measure respiration rate), etc.

Sensors can be classified into two categories based on power supply requirement: active sensors and passive sensors. Active Sensors require power supply (i.e. they provide their own energy source) and are placed on the human subject or in his surroundings. These sensors transmit and receive generated signals. They are suitable for applications in well controlled environments. Laser rangefinder is an example of active sensors. Passive Sensors do not require power supply. They deal with natural signal sources such as visual light, require no wearable devices, and only detect the transmitted energy. They are useful for surveillance applications but they can be used for all applications. Optical cameras are examples of passive sensing.

**3-3 Motion Taxonomies**

Understanding human motion requires its classification into various levels of abstractions or details. Different taxonomies that categorize motions into different levels already exist in the literature; however, some terms (such as action, activity, simple action, complex action, etc) have different meanings in different taxonomies. Here, we will review some of these taxonomies as follows:

Nagel [47] classified motion into five levels: change, event, verb, episode, and history, where a change refers to a discernable motion in a sequence, an event is a change that is considered as a primitive of a more complex description, a verb describes some activity, an episode describes a complex motion that may be consisted of several actions, and a history which is an extended sequence of related activities. Nagel's goal was to generate conceptual descriptions of image sequences. He used this taxonomy to reflect different dimensions of the motion understanding problem.

Bobick [48] used another taxonomy: movement, activity, and action where movements are the most atomic primitives requiring no contextual or sequence knowledge to be recognized, activity refers to a sequence of movements where the only required knowledge is the statistic of the sequence, and actions are larger scale events that typically include interactions with the environment and causal relationships.

Moeslund et al. [11] used the following action hierarchy: action/motor primitives, actions and activities. Action/motor primitive is an atomic movement that can be described at the limb level such as moving a leg. Action is a sequence of action primitives that may describe a possibly cyclic whole-body movement such as running. Activity consists of a set of actions that gives an interpretation of what is being performed (may be described as an understanding of the situation) such as playing football. Thus, activities are larger scale events that typically depend on the context of the environment, objects, or interacting humans. Other surveys that use the same taxonomy are [14, 19]. Turaga et al. [16] present a very similar taxonomy to that used by Moeslund et al. [11]. They used the taxonomy of atomic or primitive actions, actions, and activities. Atomic or primitive action is the simplest of action classes. Action refers to simple motion patterns usually performed by a single person and last for a short period of time (e.g., bending, walking, etc). Activities refer to complex sequences of actions performed by several humans who could be interacting with each other in a constrained manner. Activities last for much longer durations (e.g., a gang of robbers attacking a bank). The authors added, in this taxonomy, that the boundary between action and activities is not hard and that there may be some motions that lie in this grey area, where they can neither be described as simple as an "action" nor as complex as an "activity" such as of a music conductor conducting an orchestra using his gestures. Chaquet et al. [35] followed Moeslund et al. [11] and Turaga et al. [16] in structuring their taxonomy into primitive actions, actions, and activities where action is used to fulfill a simple purpose such as walking, or kicking a ball, and activity is defined as a sequence of actions over space and time such as playing football. They also related interactions as an additional feature of activities and indicated that sometimes there is no clear distinction between action and activities.

Aggarwal and Ryoo [21] categorized human motion, depending on the complexity of the motion itself, into four categories: gestures, actions, interactions, and group activities. Gestures are elementary or atomic movements performed by a part of a human body, e.g., moving a leg. Actions are activities performed by a single person and may be composed of multiple gestures, e.g., walking. Interactions are activities that involve two or more persons and/or objects, e.g., two persons fighting each other, a man shoot another one with a gun. Group activities are activities performed by conceptual groups composed of multiple persons and/or objects, e.g., a group of persons marching, two teams playing football.

Lavee et al. [17] presented a very different taxonomy that is called "event terminology". They defined an event as "an occurrence of interest in a video sequence". Inspired by some other researchers, they used prefixes to the term "event" to describe different types of events with varying properties. They used atomic and composite prefixes to describe the composition property, pixel-based and object-based prefixes to reflect the content properties, single-threaded and multi-threaded prefixes to reflect temporal properties, and sub and super prefixes to reflect the relation to the event of interest. They also introduced another term "event domain" to address the context issue by providing a description of the type of the target events, e.g., gestures in an interactive environment.

Cedras and Shah [1] considered that the recognition of higher level movements, like walking or running, should take into account that those movements consist of a complex and coordinated series of events. They defined motion events as significant changes or discontinuities in motion (e.g., a stop, a pause, a sudden change in direction or in speed, etc).

Motion can also be classified according to its type. For example, Kambhamettu et al. [49] developed a taxonomy for various types of objects motions based on the degree of nonrigidity of the object, see fig. 9. A brief review of this taxonomy is described as follows:

- Rigid motion has all distances and angles unchanged.
- Quasi-rigid motion has a small deformation; a general motion is quasi-rigid if viewed in a sufficiently short interval of time.
- Articulated motion is a piecewise rigid motion. The overall motion of the object is not rigid but its constituent parts conform to the constraints of the rigid motion.
- Isometric motion is a nonrigid motion that preserves the angles between the curves on the surface and the distances along the surface.
- Conformal motion is a nonrigid motion that preserves the angles between the curves on the surface, but not the distances.
- Homothetic motion is a nonrigid motion with a uniform expansion or contraction of the surface.
- Elastic motion is a nonrigid motion that preserves some degree of continuity or smoothness.

- Fluid motion is a nonrigid motion that violates even the continuity assumption. It may involve topological variations and turbulent deformations.

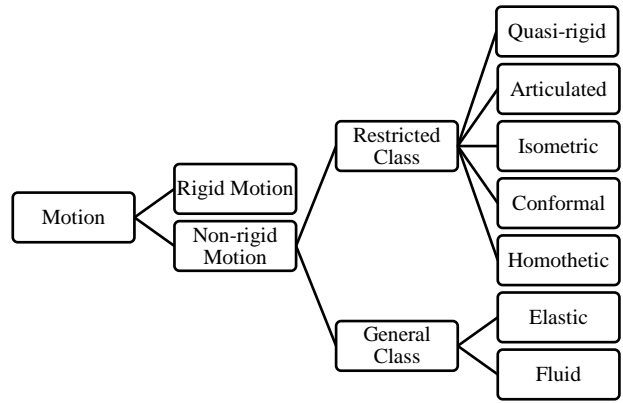

Fig. 9 The classification tree for various types of objects motions as defined by Kambhamettu et al.

**3-4 Different Stages of the Problem**

In this subsection, we will, briefly, review some classifications of various processes involved in human motion analysis and behavior understanding that are used throughout the literature.

Wang et al. [6] grouped processes involved in human motion analysis in three levels. The low level deals with human detection and contains motion segmentation and object classification processes. The intermediate level deals with human tracking. The high level deals with behavior understanding and contains action recognition and semantic description.

Aggarwal and Park [9] categorized processes involved in human activities understanding into two levels: low–level vision processes such as segmentation, tracking, pose recovery, and trajectory estimation; high-level vision processes such as body modeling and action representation.

Hu et al. [10] used a three-level hierarchy for classification of different processes according to the general framework of visual surveillance: low-level vision, intermediate-level vision, and high-level vision. The hierarchy starts with environment modeling, motion segmentation, and object classification, then continues with object tracking and ends with behavior understanding and person identification.

Turaga et al. [16] categorized processes involved in real-life activity recognition systems into three levels. At the lower levels, there are the modules of background–foreground segmentation, tracking and object detection; the main challenge of this level is to achieve robustness against errors. At the midlevel, there are action–recognition modules; the main challenge of this level is to achieve view and rate-invariant representations. At the high level, there are the reasoning engines that encode the activity semantics based on the lower level action primitives; the main challenge of this level is to achieve effective semantic representations of human activities.

Cristani et al. [25] categorized background subtraction/object segmentation and object detection into the low-level stages; while grouping object tracking and activity analysis into the high-level stages.

Shah [46] used a three-level hierarchy for automatically understanding human behavior from motion imagery. The first level deals with the extraction of relevant visual information from a video sequence. The second level represents that information in a suitable form. The third level concerns with the interpretation of the represented information for the purpose of understanding and recognizing human behavior.

Chellappa and Chowdhury [50] categorized the processes into three levels with vaguely defined boundaries: low-level processes such as extracting features, segmenting regions and tracking feature over a sequence of frames, intermediate-level processes such as grouping features, depth estimation, and motion and structure estimation, high-level processes such as description of objects and scenes.

**3-5 Different Taxonomies Used for Structuring Different Surveys**

There are several factors that are used to classify previous work in human motion analysis and understanding, these factors include: model-based vs. non-model based, explicit vs. implicit shape modeling, model types (e.g., stick figures, volumetric models, surface models, etc), human body parts involved in motion analysis, full-body motion or body parts motion, human motion modeling, level of detail needed to understand human actions, space dimensionality (2D approaches vs. 3D approaches), sensor modality (e.g., visible light, infrared light, structured light, etc), sensor multiplicity (monocular vs. stereo), sensor placement (centralized vs. distributed), sensor mobility (mobile vs. stationary), active sensing vs. passive sensing, marker-based or marker-free systems, tracking one person or multiple persons, various motion-types assumptions (e.g., rigid, articulated, elastic, etc), functionality (initialization, tracking, pose estimation, and movement recognition), image representation (global representations, local representations), view-invariant action representation and recognition, spatial and temporal structure of actions, human-object interactions and group activities, pose representation and estimation (3D model-based, 3D model-free, example-based), spatial action representations (body models, image models, spatial statistics), action classification (direct classification, temporal state-space models), action recognition (static vs. dynamic recognition, or direct recognition vs. recognition by reconstruction vs. hybrid approaches, or template matching approaches vs. state-space approaches, or nonparametric approaches vs. parametric methods vs. volumetric approaches, or single-layered approaches vs. hierarchical approaches, etc), complex activities recognition and behavior understanding (graphical models vs. syntactic approaches vs. knowledge and logic-based approaches).

**3-6 Various Approaches and Algorithms Used for Implementing Different Stages of the Problem**

Many surveys have been written in the domain of human motion analysis, recognition, and understanding. Each one has its own focus and taxonomy to compare different publications. We will briefly review some of these surveys [1-26] to reveal various approaches and algorithms used for implementing different stages of the problem and to review their pros, cons, and suitability for different situations. The surveys, which mainly focus on vision-based approaches, are listed in a chronological order to reveal the progress of the field as follows:

Cedras and Shah [1] provided a review of motion-based recognition approaches prior to 1995, where they first discussed two theories about the interpretation of motion: one of them indicates that motion information is used to reconstruct a 3D structure of the moving object and then uses the structure for recognizing the motion (Structure From Motion "**SFM**"); the other theory employs motion information directly to recognize a motion without structure recovery. Next, the authors identified two main steps involved in motion-based recognition. The first step deals with the extraction of motion information and its representation. They declared that the extracted information may be represented by trajectories (e.g. velocity, speed and direction, joint angles, relative motion, spatiotemporal curves), regions of interest (e.g., mesh feature [where a binarized frame is divided into a grid, and the ratios of black to white pixels in each block is calculated, then the ordered set of ratios for each frame is called the mesh feature]), and optical flow. The second step concerns with the matching of unknown inputs with constructed models (the classification/ recognition stage). Then, the authors reviewed examples of some matching and classification methods such as scale-space of trajectories, clustering techniques (often used with features that take the form of a vector), probabilistic methods, and connectionist methods (where the detection of a feature or an event triggers one or more units at higher layers and so on until one motion model at the highest level, representing the ordered sequence, can be selected). Next, they indicated that only few publications encountered in the literature have described the whole process of motion recognition (i.e., from the input sequence till object or motion recognition) and that most of the literature has only described various representations, without discussing issues like creating a database of models and indexing it for recognition or classification. The authors also discuss recognition methods involved in cyclic motion detection and recognition, lipreading, and hand gestures interpretation. In the end, they discuss tracking and recognition of human motion (e.g., walking, running), declaring that the tracking methods can distinguish between allowed and non-allowed human body configurations through constraints implied by human body (e.g., stick figure models, volumetric models) and motion modeling (e.g., joint angles [which are more formally expressed as flexion/extension, abduction/adduction and rotation angles], key frame sequence).

Aggarwal et al. [2] provided an overview of articulated and elastic motion analysis prior to 1996. They discussed approaches used for recovering the 3D structure and motion of objects in a bottom-up strategy. These approaches are classified into two categories: model-based approaches and model-free (i.e., without a priori shape models) approaches. Model-based approaches have the advantage of simplifying the correspondence problem due to incorporating the shape characteristics that result in reducing the search space. Shape models in articulated motion analysis (e.g., human body motion) are categorized into 2D stick figures and 3D volumetric models (e.g., generalized cylinders). Shape models in elastic motion analysis (e.g., deformable objects) are classified into parametric surface models and physically-based models. Parametric models representation schemes include: polynomials, generalized cylinders, spherical harmonics, splines, superquadrics, Fourier decomposition, and implicit algebraic surfaces. Physically based models include: snakes, deformable templates, deformable superquadrics, and modal models. Many of the above deformable shape models are used for modeling human body; however, Pan et al. [34] provided a recent review about deformable human body models used in motion analysis, where they surveyed various deformable modeling methods used in the past 30 years. They categorized these methods into 2D models, 3D surface models, geometric-based, physics-based, anatomy-based, and motion-based approaches.

Gavrila [3] provided a survey on human motion analysis including both hand gesture (hand motion) and whole-body motion. The author first discussed three approaches used for tracking people. These approaches are: 2D approaches with and without explicit shapes and 3D approaches. The author declared that most of the 2D and 3D approaches, till 1997, deals only with incremental pose estimation without providing ways for bootstrapping at initialization or when tracking gets lost. Next, he reviewed different action recognition techniques (e.g., detecting periodic action using spatiotemporal templates [53], dynamic time warping "**DTW**", Hidden Markov Models "**HMMs**"). Finally, the author presented a comparison between 2D and 3D approaches explaining the applicability, design methodology, and challenges of each approach.

Aggarwal and Cai [4] discussed three major areas related to human motion analysis, which are: motion analysis of human body parts, tracking a human object from a single and multiple views, and recognizing human actions. Human action recognition approaches are categorized into: template matching and state space approaches. Template-based approaches are characterized with their lower computational cost, but they are also more sensitive to the variation of the action duration. On the other hand, state space approaches define each static posture of a human body as a state. Connections between these states represent the transitions probabilities. An action can be represented by a tour through these states. The problem of action duration variation can be treated by allowing any state to repeatedly visit itself; however, there should be a careful selection of the number of states and the dimension of the feature vector in order to avoid over-fitting or under-fitting.

Moeslund and Granum [5] provided a comprehensive survey on human motion capture using 130 papers published during the last two decades in the twentieth century with much more focus on the period (1994-2000). They used a taxonomy based on the system functionalities. Their taxonomy consists of four processes: initialization, tracking, pose estimation, and recognition. Initialization can be performed offline prior to the system operation or as the first stage of it. It mainly concerns with online or offline camera calibration, adaptation to scene characteristics (which is related to the appearance assumptions and the segmentation methods, e.g., background subtraction uses offline adaptation, temporal differencing uses online adaptation), and model initialization (which is concerned with the subject's model and its initial pose). Tracking is consisted of three stages: segmentation, representation of the segmented objects, and tracking over time. Figure-ground segmentation uses either temporal (e.g., subtraction and flow) or spatial information (e.g., threshold and statistics) of the image sequence. Subtraction approaches refer to subtracting the intensity or gradient values of the current image pixels from their previous values while flow approaches describe the coherent motion of pixels or features between the video frames. Thresholding approaches (e.g., chroma-keying where it can be easy to segment a subject in front of a one-color background, usually blue, wearing nonblue clothes) are suitable for controlled environments with restricted appearance assumptions, while statistical methods, which deal with the characteristics of pixels or group of pixels to extract figures from the background, are a much better choice for unconstrained applications due to their ability of adaptability. Representations of the segmented entities can be classified into: object-based representation that depends on the segmentation output (e.g., points, blobs, boxes, silhouettes) and image-based representation that is derived directly from the image (e.g., spatial, spatio-temporal, edges, features). Tracking over time is to find the object correspondences in consecutive frames. This can be eased by prediction. Kalman filter and the condensation algorithms are common methods for predictions. Pose estimation is discussed here according to how various systems apply human models (model free, indirect model use, direct model use). Model free approaches mean there is no a priori model used in the system. Indirect model use means that the model is used as a guide for the interpretation of the measured data. Direct model use means that the model is continuously updated and therefore it includes an estimate of the pose at any time. Human motion recognition can be classified into recognition by construction and direct recognition. Another distinction among recognition approaches may be based on one or more frames (i.e., static or dynamic recognition). Static recognition deals with spatial information one frame at a time by comparing prestored information with the current image to recognize postures while dynamic recognition incorporates temporal characteristics over the image sequence.

Wang et al. [6] provided a comprehensive survey on three major issues involved in a general human motion analysis system, which are human detection, tracking, and activity understanding. The survey covers the research published from 1989 to 2001 with nearly 70% of the discussed papers were published after 1996. The detection process usually involves motion segmentation and object classification. Motion segmentation approaches include: background subtraction, statistical methods (or adaptive background modeling which is characterized with its robustness against noise, illumination change, shadows, etc), temporal differencing, and optical flow. Object classification is important for tracking the required object. Approaches for object classification include shape-based approaches (e.g., blob area and dispersedness), motion-based approaches (e.g., detecting motion periodicity by using self similarity measure techniques, calculating residual flow to detect rigidity or periodicity of moving objects), combination of shape-based and motion based approaches, and combination of component based and geometric configuration constraints of the human body. Useful mathematical tools used for tracking include: Kalman filter (deals with unimodal distributions), condensation algorithm (deals with multi-modal distributions, cluttered background), and DBNs. Tracking is divided into four categories: model-based, region-based, active contour-based, and feature-based. Model-based include: stick figures (which may be obtained by median axis transform), 2D contour (which resulted from body projection to the image plane), volumetric models (such as elliptical cylinders, cones, spheres, etc). Active contour models or snakes approach dynamically updates the bounding contour of the human object (it requires a good initial fit but it reduces computational complexity compared to region-based approach. It also can deal with partial occlusion). Feature-based tracking includes feature extraction and feature matching (e.g., corner points of moving silhouettes can be selected as features and a distance measure based on positions and curvatures of points can be used for the matching stage). There is a tradeoff between feature complexity and tracking efficiency. Behavior Understanding involves both action recognition (template matching and state-space approaches) and semantic descriptions (choosing the proper motion words or short expressions to describe the behavior of moving objects). It may be viewed as a classification problem of time varying feature data (i.e., matching a test sequence with reference ones).

Buxton [7] discussed generative models used in learning and understanding dynamic scene activity. First, the author pointed to explicit and exemplar models before handling flexible generative models. Then, she indicated how some flexible generative models can be regarded as extensions of each other declaring that more capability comes at the cost of more complexity. Next, two types of generative models were studied: deformable models and graphical models, indicating their complementary nature as deformable models have been used to represent static and dynamic shape, texture and other physically observable parameters and that more general graphical models have been used to capture more abstract relationships. Point distribution model "**PDM**" (which is based on a set of example shapes of a given object where each shape is defined by landmark points that are corresponding to the important object features) was discussed as an example of deformable models. For graphical models, several examples were given such as Bayesian belief networks "**BBN**", dynamic Bayesian networks "**DBN**", dynamic decision networks "**DDN**" and HMMs. Moreover, the author referred to the use of CHMM for modeling interactions, VLHMM for interpreting gestures, and parameterized HMM for virtual reality systems. Finally, a discussion about the use of generative models in applications such as smart rooms and visual surveillance was provided.

Wang and Singh [8] provided a survey on two main areas of human dynamics analysis that are related to the biometric research. The first area concerns with the tracking of the human body as a whole and the tracking of individual human body parts such as face, head, hands, fingers. The second one concerns with modeling behavior using full body motion analysis, hand motion analysis (gesture analysis), and leg motion analysis (gait analysis).

Aggarwal and Park [9] discussed four aspects of high-level processing involved in modeling and recognition of human actions and interactions. These aspects are: human body modeling, abstraction details, human action recognition approaches, and domain knowledge involved in high-level recognition schemes. The authors classified approaches used to study articulated human motion into two categories: model-based (which use a prior shape model) and appearance-based (no shape model is used). Model-based approaches can deal more efficiently with complex motions because of its ability to integrate shape knowledge and visual input; however, they require more processing to match the model with the input image. Appearance-based approaches can be applied in more applications, but they are sensitive to noise. Human body modeling varies, according to the application, from coarse representation such as bounding box to fine representation such as 3D volumes. Recognizing human actions and interactions can be performed at different abstraction levels: gross (e.g., human body may be represented by bounding box), intermediate (e.g., human body may be represented by its major part such as head, torso, etc), and detailed level (a single body part is used to recognize the performed action). For example, the gross level may be suitable for surveillance applications whereas the detailed level is suitable for human computer interaction. The authors reviewed some examples of interactions between a human and objects (e.g., recognizing human actions, such as opening a cabinet and picking up a phone, in a static room using a system that utilizes a scene context [54]) and between two persons (e.g., recognizing interactions such as following another person or walking together by using a statistical Bayesian system that combines both top-down and bottom-up approaches [55]). Then, they discussed two classifications of human action recognition approaches. The first one classifies approaches into: direct recognition (i.e., recognizing human actions directly without the reconstruction of body part poses by using, e.g., periodicity information in cyclic motion to recognize walking, jumping jack, etc [56]), recognition by reconstruction (i.e., constructing body poses before recognizing actions), and hybrid approaches. The second classification is static vs. dynamic recognition. Static recognition utilizes static representations of individual frames. Here, most approaches apply template matching for recognition. Dynamic recognition utilizes dynamic representation of the entire sequence or a part of it. Most approaches of this kind apply DTW or HMM. Finally, the authors reviewed some interpretation schemes such as the rule-based network, physical constraints, casual analysis, syntactic analysis, and finite automata methods.

Hu et al. [10] discussed different stages involved in the visual surveillance system. These stages are environment modeling, motion detection, object classification, tracking, understanding and description of behaviors, human identification, and data fusion from multiple cameras. The authors extended their previous survey [6] by elaborating more areas and discussing other related issues such as occlusion handling, fusion of 2D and 3D tracking, and remote surveillance.

Moeslund et al. [11] provided a very comprehensive survey by reviewing over 350 publications reflecting the advances in human motion capture and analysis from 2000 to 2006. They followed the taxonomy of the previous survey of Moeslund and Granum [5]. The authors indicated the role of surveillance applications to push the progress in the field especially in the areas of tracking and human action recognition. They discussed how the research has addressed novel methodologies for achieving automatic initialization, reliable tracking (e.g., tracking through occlusion) and pose estimation in natural scenes, and automatic recognition of simple human actions and behaviors. The authors also indicated that only a small body of the literature has addressed issues like complex activity recognition (such as interactions with other people), or the use of scene contextual information to interpret actions.

Yilmaz et al. [12] provided a comprehensive survey where they categorized different tracking methods based on object and motion representations, discussed representative methods in each category along with their pros and cons, and provided qualitative comparisons of these different tracking algorithms. Moreover, they discussed other important issues related to tracking such as object representation, appropriate features selection, and object detection. They first categorized object representations according to their shape and appearance. Object shape may be represented by its centroid or a set of points (which is suitable for objects occupying small image areas), primitive geometric shapes (e.g., rectangle, ellipse), contour, silhouette, an articulated shape model, or a skeletal model. Object appearance may be represented by its probability density function, which can be either parametric (such as Gaussian and a mixture of Gaussians), or nonparametric (such as histograms and Parzen windows). Other appearance representations include templates, active appearance models, and multi-view appearance models.  Multi-view appearance models can be computed by subspace approaches (such as principal component analysis "**PCA**"**,** Independent component analysis "**ICA**"**),** or by training classifiers (such as support vector machines "**SVM**" and Bayesian Networks "**BN**s"). Next, the authors discussed some features that are used to characterize objects such as color, edges, optical flow and texture. They categorized automatic feature selection into filter methods (e.g., PCA) and wrapper methods (e.g., adaptive boosting). The authors indicated that there are many tracking algorithms which employ a weighted combination of different features in order to make an object more discriminative from other objects in the scene and from the background. Tracking methods usually require the detection of objects either in every frame (e.g., point trackers) or when they first appear in the image sequence (e.g. geometric region or contour-based trackers). Objects can be detected by using: point detectors (e.g., Moravec's detector, Harris detector, Scale Invariant feature transform "**SIFT**", Affine Invariant Point detector), background subtraction (e.g., dynamic texture background), image segmentation into perceptually harmonic regions (e.g., mean shift, graph cut, active contours), and supervised learning of different views of an object (e.g., SVM, neural networks, Adaboost). Next, the authors discussed various object tracking algorithms. They categorized different trackers into point trackers, kernel trackers, and silhouette trackers. Point correspondence trackers are categorized into deterministic methods (e.g., multi-frame point correspondence tracker using a non-iterative greedy algorithm "**MFT**") and statistical methods (which include approaches for single object [e.g., Kalman filters and particle filters] and approaches for multiple objects [e.g., joint probability data association filter "**JPDAF**", multiple hypothesis tracking "**MHT**", probabilistic MHT "**PMHT**"]). Kernel trackers are categorized based on the used appearance representation into templates and density-based appearance models (e.g. mean shift, Kanade-Lucas-Tomasi tracker "**KLT**", layering approach), and multiview appearance models (which include subspace approaches [e.g., eigentracking] and classifier approaches [e.g., SVM]). Silhouette trackers are categorized into shape matching trackers (e.g., Hough transform, Hausdorff distance, histograms of color and edges) and contour-based trackers. Contour-based trackers, in turn, are categorized into state space models and methods that utilize direct minimization of contour energy function through using either variational or heuristic approaches. Finally, the authors discussed other issues such as occlusion handling and multiple camera tracking.

Poppe [13] discussed the characteristics of human motion analysis by studying pose estimation process in model-based (or generative) approaches and model-free (or discriminative) approaches. For model-based approaches, the pose estimation process consists of two phases: a modeling phase and an estimation phase. The modeling phase is the construction of a likelihood function, that takes into account: the human body model (which describes both the kinematic and shape properties of the human body), image descriptors (which include silhouettes and contours, color and texture, edges, 3D reconstruction created from silhouettes, optical flow motion, and a combination of these descriptors), camera considerations (e.g., single camera or multiple cameras, data fusion or active viewpoint selection), and environment settings (e.g., indoors or outdoors, single person or multiple persons). Matching functions between the visual inputs and the generated appearance of the human models are also discussed according to the used image descriptors. The estimation phase searches for the most likely pose given the likelihood surface. Three approaches for model-based estimation are discussed, which are top-down estimation, bottom-up estimation, and a combination of these two approaches. Other issues concerns with the estimation phase such as single and multiple hypothesis tracking, 3D estimation from 2D points, motion models, and pose space dimensionality reduction are also discussed. For model-free approaches, the pose estimation process can be accomplished through learning-based approaches (using extensive training data for learning a mapping function from image space to pose space) or example-based approaches (building a large database of exemplars and their pose descriptions, here a similarity search is performed and candidate poses are deformed or interpolated to obtain the most likely pose that match the image observation).

Krüger et al. [14] reviewed different approaches for the representation, recognition, understanding, and synthesis of actions within the computer vision, robotics and artificial intelligence communities. They first categorized the approaches used to recognize human actions into four categories: scene interpretation (activities are recognized without knowing the identity of the moving objects), full-body-based (approaches here do not use explicit human models but rather depend on 2D image projections, e.g., space-time volumes [**XYT** volumes], direct pattern recognition such as PCA, MEI, MHI, timed MHI "**tMHI**", local linear embedding "**LLE**", etc), body-part-based, and action primitives and grammars. Then, they reviewed representative examples in each category. Next, they discussed action learning and imitations indicating that robotics, unlike vision, concerns mainly with generative models of actions to enable imitation learning, but in cases where pure recognition is the only requirement in the context of understanding or interaction, different discriminate approaches are commonly used. They also indicated that the discovery of mirror neurons has greatly influenced many researchers in robot imitation (e.g., some researchers propose a HMM model in which movement primitives can be both recognized and generated, thus realizing the idea of mirror neurons on a humanoid robot). In the end, they reviewed many approaches used for plan or intention recognition such as rule-based systems, Bayes nets, parsing of context-free grammars, graph covering, etc.

Pantic et al. [15] indicated that in order to enable computers to understand human behavior, human computer interaction (**HCI**) designs must be switched from computer-centered designs towards human-centered designs. They indicated that this shift requires the design of human-like interactive functions such as understanding and imitating certain kinds of human behaviors like affective and social signals. Then, they discussed three issues required to design these function, which are: what is communicated, how the information is passed, and in which context the information is passed on. For the first requirement, they discussed five types of messages conveyed by behavioral signals, which are affective/attitudinal states, manipulators, emblems, illustrators, and regulators. For the second requirement, they discussed facial expressions, body gestures and nonlinguistic vocalizations modalities and indicated that combining multiple modalities may be useful for the realization of automatic human behavior. For the third requirement, they discussed six key aspects of the contextual information (or the so-called W5+: who is the observed user, where is the user, what is the current task of the user, when does the user perform the behavioral signals, why does the user perform these signals, and how does the user perform these signals). The authors also discussed a number of tasks involved in modeling human behavior and understanding displayed patterns of behavioral signals (e.g., sensing and analyzing human behavioral signals, context sensing, and interpretation of the human behavioral signals and context information into a description of the shown behavior). They also indicated that most of the machine learning approaches of human behavior till 2007 is neither multimodal, nor context-sensitive, nor suitable for handling longer time scales, so they urged the researchers to address these issues and treat the problem of context-constrained analysis of multimodal human behavioral signals as a one complex problem rather than a number of isolated problems. In the end, they pointed out a number of scientific and technical challenges to further the progress in the field. Scientific challenges include: handling too much information from different behavioral channels (e.g., face, body, vocal nonlinguistic signals) may be confusing, data fusion, how the grammar of human behavioral displays can be learned, building systems that can be educated through experience. Technical challenges include: automatic initialization, robustness, real time processing, developing benchmarks for comparing different approaches and evaluate the progress.

Turaga et al. [16] presented a comprehensive survey on representing, recognizing, and learning human actions and activities from video and related applications. They discussed the problems at different levels of complexity starting from atomic or primitive actions, where they briefly discussed low-level feature extraction using optical flow, point trajectories, blob and shape segmentation, and filter responses techniques. Three shape descriptors are discussed: global (the entire shape is considered), boundary (only the shape contour is considered) and skeletal descriptors. Spatiotemporal filter responses are used to classify several features, and they are more suitable for low resolution and poor quality videos than methods like optical flow. Next, the authors dealt with actions with more complex dynamics, where they categorized the used approaches into three major classes: nonparametric, volumetric and parametric approaches. Nonparametric approaches extract a set of features from each frame of the video and compare it to stored templates. Nonparametric approaches contain 2D templates and 3D models. Volumetric approaches consider a video as a 3D volume of pixel intensities and are categorized into: spatio-temporal filtering, part-based approaches, subvolume matching and tensor-based approaches. Parametric approaches deal with the temporal dynamics of motion. Examples of these approaches include HMMs and their variants (e.g., coupled HMM, mixed HMM. semi HMM), linear dynamic systems "**LDS**" (which are considered as continuous state-space generalizations of HMMs) and nonlinear dynamic systems (such as switching linear dynamic systems "**SLDS**" and jump linear systems "**JLS**"). Next, the authors considered complex activities, where they categorized various approaches into three categories: graphical models, syntactic approaches, and knowledge and logic-based approaches. Graphical models include DBNs, Petri nets (a mathematical tool used for describing the relations between conditions and events), propagation nets, past-now-future "**PNF**" network, suffix trees, etc. Syntactic approaches include grammars, attribute grammars, and stochastic grammars. Knowledge and logic-based approaches include constraint satisfaction, logic rules, and ontologies. Finally, the authors discussed invariance in human action analysis by reviewing different approaches used to deal with invariance in viewpoint, execution rate and anthropometry.

Lavee et al. [17] discussed two main components of the event understanding process (i.e., the process that translates low-level content in video sequences into high level semantic concepts). These two main components are: abstraction and event modeling. Abstraction is defined as the process of molding the data into informative units or constructs (sometimes called "primitives") to represent the abstract properties of the video sequence and to be used as input to the event modeling stage (e.g., what features to be selected). The authors indicated three types of abstraction which are: pixel-based (e.g., color, texture), object-based (e.g., bounding boxes, silhouettes), and logic-based (i.e., semantic rules and concepts). Event modeling concerns with the representation and recognition of events as they occur in the video sequence. The authors indicated three types of classifications that event modeling may be classified into: deterministic versus probabilistic, generative versus discriminative, or pattern-recognition methods versus state models versus semantic models. The authors used the third classification type to discuss event models in their survey. Examples of pattern-recognition methods include SVM, nearest neighbor, neural networks, and boosting. State modeling include finite-state machine "**FSM**", BNs, DBNs, HMMs and their variants (coupled HMMs "**CHMMs**", layered HMMs "**LHMMs**", multiobservation HMMs "**MOHMMs**", dynamic multi-level HMMs "**DML-HMMs**", parallel HMMs "**PaHMMs**", hierarchical HMMs "**HHMMs**", hierarchical semi parallel HMMs "**HSPaHMMs**", variable length Markov models "**VLMMs**", hidden semi-Markov models "**HSMMs**", coupled hidden semi-Markov models

"**CHSMMs**", switching hidden semi-Markov models "**S-HSMMs**"), and conditional random fields "**CRFs**". The authors also discussed semantic models such as grammar models, Petri nets "**PNs**", constraint satisfaction and logic-based approaches. In the end, they included a comparison table of 43 publications indicating both the used abstraction scheme and the used event model, besides examples of the modeled events (e.g. walking, running, etc).

Ji and Liu [18] provided a survey on view-invariant human motion analysis with the emphasis on view-invariant pose representation and estimation, and view-invariant action representation and recognition. First, they considered human detection by briefly indicating some conventional approaches used in motion segmentation (background subtraction, temporal differencing, and optical flow) and object classification (shape-based classifier, motion-based classifier, and a hybrid classifier utilizing both of them). Next, they categorized view-invariant pose representation and estimation into three categories, which are: 3D model-based, model-free, and example-based. 3D model-based approaches represent both human shape and kinematics geometrically in 3D and then optimize the similarity between the model projections and the input images. The authors discussed these approaches considering the following points: dimensionality reduction, used estimation techniques (e.g., Kalman filter, condensation algorithm and DBNs), sub categorization into single-view and multiple-view, and applicability in constrained settings. Model-free approaches involve 3D reconstruction of both human shape and motion directly from a visual hull (also known as shape from silhouette "**SFS**") without a prior model. Example-based approaches store a database of human motion figures and then a similarity measure (e.g., parameter-sensitive hashing, chamfer distance, enhanced pyramid match kernel algorithm "**PMK**") is conducted between candidate examples and the input video frame to estimate the pose. The authors also reviewed three methods based on normalization where the view-point is removed by transforming all observations into a canonical coordinate frame (e.g., projecting the video frames onto the ground plane, synthesizing a side view using perspective projection model or optical flow based structure from motion). Next, they categorized behavior understanding into two categories: action recognition and behavior description. Then, they reviewed view-invariant action representation and recognition approaches and categorized them into template-based approaches (e.g., MHVs [57], STVs [58], 2D spatiotemporal curvature [59], space-time shapes [60], geometric invariants computed from five points that lie in a plane [61]) and state space approaches (HMMs, CRFs, Action Net) advising to use CRFs instead of HMMs to deal with view-invariant human motion analysis because of their ability of modeling dependencies between features and observations. In the end, they reviewed few approaches for behavior description (e.g., probabilistic context-free grammar "**PCFG**", stochastic context-free grammar "**SCFG**").

Poppe [19] discussed the vision-based human action recognition as a combination of feature extraction and subsequent classification of these image representations. Therefore, he first discussed global and local image representations. Global representations are acquired in a top-down fashion where a person is detected using a background subtraction technique or a tracking algorithm, and then the detected region (e.g., silhouettes) is encoded as a whole. Local representations are calculated in a bottom-up fashion where spatiotemporal interest points are detected first, then local patches around these points are calculated, and finally theses patches are combined to form a final representation. Global representations are more sensitive to viewpoint, noise and occlusion but they encode more information. Finally, the author discussed human action recognition as a classification problem where he addressed three approaches: direct classification approaches (i.e. approaches that do not explicitly model variations in time), temporal state-space models (which model action variations in time), and action detection approaches (i.e., approaches that neither model the object explicitly nor the action dynamics).

Weinland et al. [20] reviewed and categorized different approaches used in action representation, segmentation and recognition, concentrating on full-body motions, such as kicking, punching, and waving. The authors first categorized different approaches used to represent actions according to the spatial and temporal structure of actions. In the spatial domain, actions can be represented based on global image features (i.e., image model of an action, where a region of interest around the person is detected first then a dense computation of features within a grid bounded by the region of interest is performed), or parametric image features (i.e., human body features), or spatial statistical modeling (i.e., modeling the spatial distribution of local image features). In the temporal domain, actions can be represented based on global temporal features (i.e., action templates), or grammatical models, or temporal statistical models (a single characteristic key frame is used to model the appearance of an action). Next, the authors discussed different methods used for temporal segmentation of an action. They categorized theses methods into three classes, which are boundary detection, sliding windows, and grammar concatenation. Finally, the authors discussed view-independent action recognition after categorizing them into: view normalization (i.e., mapping each observation to a common canonical coordinate frame), view invariance (search for features that are independent on the class of view transformation considered), and exhaustive search (search over all possible transformations).

Aggarwal and Ryoo [21] provided a review of human activity analysis. They discussed the methodologies of recognizing simple actions and high-level activities. They classified all activity recognition methodologies into two categories: single-layered approaches and hierarchical approaches. Single-layered approaches are used to analyze single actions of a single person. These approaches are categorized into space-time approaches (which include spatiotemporal trajectories, space-time volume, and space-time features) and sequential approaches (which include exemplar-based and state-based approaches). Hierarchical approaches are used to analyze high-level activities including those performed by multiple persons. These approaches include statistical, syntactic, and description-based approaches. The authors also discussed the recognition of human-object interactions and group activities. Moreover, they compared the advantages and disadvantages of each approach.

Chen and Khalil [22] briefly discussed vision-based and sensor-based activity recognition approaches indicating a new emerging object-based approach that deals with a sensorized environment where activities are characterized by objects being used during their operation. They also briefly discussed some activity recognition algorithms (machine leaning techniques [which include both supervised and unsupervised learning techniques that primarily use probabilistic and statistical reasoning] and logical modeling reasoning). However, the main interest of their article is given to describe the general framework and the lifecycle of the ontology-based activity recognition approach where they discussed domain knowledge acquisition by deriving relevant information from interviews and questionnaires or by using information extraction and data mining from the web, domain knowledge modeling using context and activity ontologies, generating semantic descriptions through using a generic ontology editor to manually create all semantic instances and then converting collected sensor data to textual descriptions, archiving semantic descriptions in repositories for later use by services and applications, using manually developed activity ontologies as seed ontologies to allow both the recognition of activities and their growth through adding new instances, etc. The authors also provided an exemplar case study "MakeDrink" to demonstrate the ontology-based approach indicating its support to progressive activity recognition at both coarse-grained and fined-grained levels.

Yang et al. [23] argued that most tracking methods include: video input, appearance feature descriptions, context information, decision making, and model update. They referred to different feature descriptors of an object such as: gradient features (descriptors are categorized in two groups: shape- or contour-based descriptors such as chamfer matching, and statistical descriptors such as SIFT, speeded up robust features "**SURF**", histogram of oriented gradients "**HOG**", adaptive contour feature "**ACF**", granularity tunable gradients partition "**GGP**", position orientation histogram, Hough transform, Hough forest detection), color features (histogram- and SIFT-based color descriptors), texture features (Gabor wavelet, local binary patterns "**LBP**" descriptor [which is characterized by its tolerance against illumination changes and its simple computations] and its variants including: multi-scale block **LBP**, semantic **LBP**, Fourier **LBP**, local ternary patterns "**LTP**", and local Gabor binary pattern, Weber local descriptor), spatio-temporal features (e.g., **3D SIFT**, extended SURF "**ESURF**", dynamical local binary pattern "**DLBP**"), biological features (Enhanced biologically inspired model "**EBIM**", attention regions "**ARs**") and fusion of multiple features (e.g., covariance matrix, sigma set, **HOG-LBP**, generic objectness measure). The authors also indicated that there is no single feature descriptor is robust enough to handle all situations. For example, the HOG descriptors fails to handle noisy edge regions, the color histogram fails when handling objects with different textures, etc. Finally, they categorized the tracking progresses into three groups: online learning methods (generative approaches and discriminative approaches "tracking by detection"), context information, and Monte Carlo Sampling, where they reviewed representative methods in each group.

Holte et al. [24] reviewed and compared multi-view approaches for human 3D pose estimation and activity recognition. They first dealt with model-based approaches aimed to extract 3D postures indicating common steps involved in these approaches. They pointed out that for each step, different approaches may have different choices. For example, some approaches may use 3D features such as reconstructed voxel data, or 2D features such as color, silhouettes. Another choice may be whether to use temporal information from previous frames (tracking-based) or the current frame only (single frame-based) in the mapping from the input space to the body model configuration space (i.e. estimating the body pose). Also, the pose initialization step may be set automatically or manually. A last choice may be between generic and specific approaches (the latter ones prove usefulness for efficiency). Next, the authors dealt with 2D and 3D approaches used for human action recognition. They referred to various shape and motion features and action representations used in different approaches (e.g., spin image, feature trees, shape histogram, 3D shape context, 3D histogram of oriented gradients (**3D HOG**), motion history volumes (**MHVs**), spatio temporal volumes (**STVs**), 4D action feature model (**4D-AFM**), 3D motion context (**3D-MC**), harmonic motion context (**HMC**), computing **R** transform surfaces of silhouettes and manifold learning to describe actions in a view-invariant manner, applying spherical harmonics to represent the shape histogram in a view invariant manner). They also indicated a number of methods used in the classifier/matching phase such as HMMs, SVMs, Bayesian classifiers, nearest neighbor, similarity matrix, Mahalanobis distance, Euclidean distance, Hamming distance, 2D diffusion distance, template matching, graph matching, etc. Moreover, the authors pointed out a new strategy that has been used recently by several authors to recognize actions, which is known as cross-view recognition where training is done in one view while testing is done on another completely different view (e.g., side view versus top view). In the end, they discussed some shortcomings of the multi-view camera systems (e.g., noisy foreground segmentation results in a reduced quality of the reconstructed shape from silhouette, self-occlusion problem may lead to merging body parts) and the pros and cons of using sensors like ToF range cameras to acquire 3D data (e.g., low cost, issues like the correspondence problem and careful cameras placement and calibrations are avoided, limited range of about 6-7 meters, capturing only 3D data of frontal surfaces).

Cristani et al. [25] analyzed a new perspective of human behavior analysis that brings in concepts and principles from the social, affective, and psychological literature, and that is called Social Signal Processing (**SSP**). The authors discussed, first, the use of SSP in different stages of a typical video surveillance system indicating its importance in both the dynamic phase of a tracker and the activity analysis stage. Then, they introduced a short review about classical activity analysis indicating that the use of spatio-temporal features instead of trajectories is gaining more popularity due to their finer analysis of individual human behavior and their robustness to noise, small camera motion, and lighting changes. They used two taxonomies in their short review. The first one categorizes different approaches by considering three aspects which are: degree of environmental supervision, level of detail of interaction, and number of subjects. The second taxonomy considers the type of methodologies being used for classifying them into three categories: graphical or generative models (which include Markov models, Bayesian networks, dynamic Bayesian networks (mostly HMM), generic Bayesian models, nonparametric models, etc), discriminative approaches (e.g., SVM, latent SVM, relevant SVM, SVM with different kinds of kernels, neural networks), syntactical approaches (e.g., grammars). Next, the authors provided three examples of problems that encode social events and are encountered very frequently in video surveillance, which are: definition of a threatening behavior, modeling of groups, and modeling of interactions in outdoor scenarios. They indicated that employing SSP in these problems would be apparently fruitful. Then, they discussed various behavioral cues that represent heterogeneous and multimodal aspects of a social interplay. These cues are categorized into five categories: physical appearance (e.g., attraction, height, somatotype), body postures and gestures, facial expression and gazing, vocal characteristics (prosody, linguistic and non linguistic vocalization, silence, turn taking patterns), and space and environment (interpersonal distances, spatial arrangements of interactants). They indicated that most SSP approaches deal with highly constrained environments where massive and pervasive sensors are installed such as smart rooms, while their use in smart surveillance is still in its infancy level. They also discussed crowd behavior analysis considering three main applications which are density estimation of the crowd, tracking some individuals in the crowd, and crowd behavior understanding. They also pointed out that crowds can be modeled using social force or correlation topic models. In the end, they indicated that combining sociological notions with computer vision algorithms may lead to novel applications such as design of public spaces and learning spaces.

Smeulders et al. [26] provided a recent survey about application independent trackers where they have evaluated the performance of nineteen single object trackers using the 315 video clips of the **ALOV++** dataset focusing on the accuracy and the robustness of the these trackers. The videos contain diverse situations such as occlusion, illumination changes, moving camera, zooming camera, clutter, low contrast, severe changes of object shape, nearby similar objects, and other more aspects of tracking circumstances. The selected trackers came from different origins to cover a wide range of paradigms. Half of the trackers were chosen from well-cited publications in the period from 1999 to 2006, and the other half were published in major conferences in recent years. These trackers are: Normalized Cross-Correlation "**NCC**", Lucas-Kanade Tracker "**KLT**", Kalman Appearance Tracker "**KAT**", Fragments-based Robust Tracking "**FRT**", Mean Shift Tracker "**MST**", Locally Orderless Tracker "**LOT**", Incremental Visual Tracker "**IVT**", Tracking on the Affine Group "**TAG**", Tracking by Sampling Trackers "**TST**", Tracking by Monte Carlo sampling "**TMC**", Adaptive Coupled-layer Tracker "**ACT**", L1-minimization Tracker "**L1T**", L1 Tracker with Occlusion detection "**L1O**", Foreground-Background Tracker "**FBT**", Hough-Based Tracker "**HBT**", Super Pixel tracking "**SPT**", Multiple Instance Learning Tracking "**MIT**", Tracking, Learning and Detection "**TLD**", STRuck: Structured output tracking with kernels "**STR**". All of these trackers belong to the class of online trackers where a target is selected by bounding a box around it in the first frame. No pre-training data were used (i.e. the tracked target is unknown). The authors demonstrated that the above trackers can be evaluated objectively by using survival curves, Kaplan Meier statistics and Grubbs outlier testing.

Before we conclude this subsection, we want to point out the importance of background modeling in human motion detection; we therefore refer to some good references about this issue as follows:

Toyama et al. [27] provided a list of ten canonical problems to which an ideal background modeling system should be able to solve. These canonical problems are: moved background object, gradual illumination change, sudden illumination change, waving trees, camouflage effect, bootstrapping, foreground aperture, a foreground object becomes motionless, an object starts motion, and shadows. Next, the authors developed the wallflower algorithm, a three-level system (pixel level, region level, and frame level), for adaptive background modeling. Then, they provided a comparison between ten different background modeling algorithms, including their two systems, using scenarios that typify the first seven problems. These algorithms are: adjacent frame differencing, mean and threshold, mean and covariance (Pfinder approach) [30], mixture of Gaussians [31], normalized block correlation, temporal derivative (W4) [32], Bayesian decision, eigenbackground, linear prediction and wallflower. Finally, the authors proposed a set of five principles, upon the analysis of the experimental results, to which a background modeling algorithm should stick to. These principles are: avoiding semantic differentiation of foreground objects, proper initial segmentation, adaptation to both sudden and gradual changes in the background, defining an appropriate pixel-level stationarity criterion, and modeling changes at different spatial scales.

Piccardi [28] provided a review on seven background subtraction techniques, which are: running Gaussian average (Pfinder approach) [30], temporal median filter, mixture of Gaussians [31], kernel density estimation "**KDE**", sequential kernel density approximation "**SKDA**", cooccurence of image variations, and eigenbackground. Then, the author provided a comparative performance analysis of these techniques based on speed, accuracy, and memory requirements in order to enable the system designer to choose the most suitable technique for a specific application. The analysis shows that the running Gaussian average and the median filter techniques can achieve acceptable accuracy with limited memory requirements while processing a high frame rate sequence. Mixture of Gaussians and KDE provide a very good accuracy but KDE requires a large amount of memory. SKDA, which is an approximation of KDE, can almost achieve the same accuracy of KDE but requires lower memory and time complexities. Eigenbackground and cooccurence of image variations techniques, which explicitly address spatial correlation, can provide good accuracy while having reasonable time and memory complexities, but the last technique imposes a tradeoff with resolution.

Hall et al. [28] compared the target detection performances of five adaptive background differencing methods. These methods are: basic background subtraction "**BBS**", W4 [32], Single Gaussian Model "**SGM**" (Pfinder approach) [30], mixture of Gaussian **MGM** [31], and LOTS. The performances of these methods are evaluated on the CAVIAR entry hall sequences of the PETS 2004 dataset with manually annotated ground truth data. These sequences are characterized by difficult lighting conditions in indoor scenes. The authors found that the methods LOTs and SGM produce better results than the more complex background model MGM.

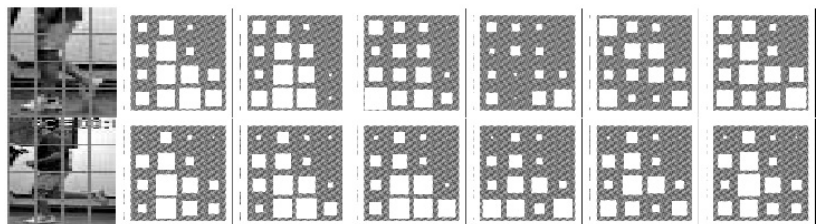

(a)

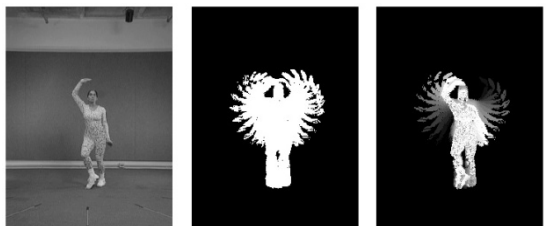

(b)

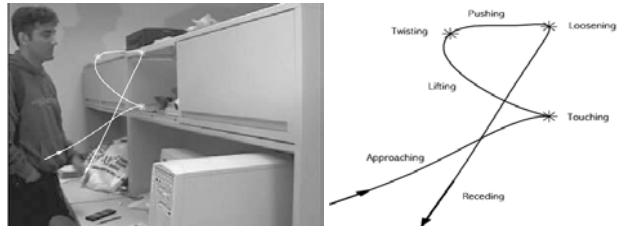

(c)

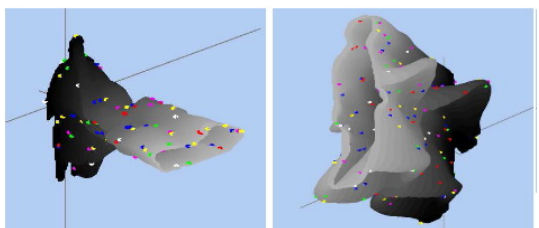

(d)

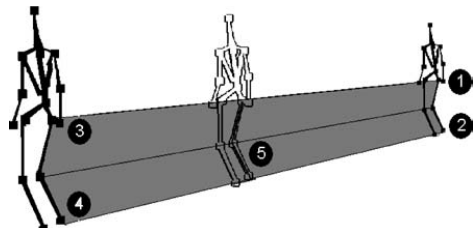

(e)

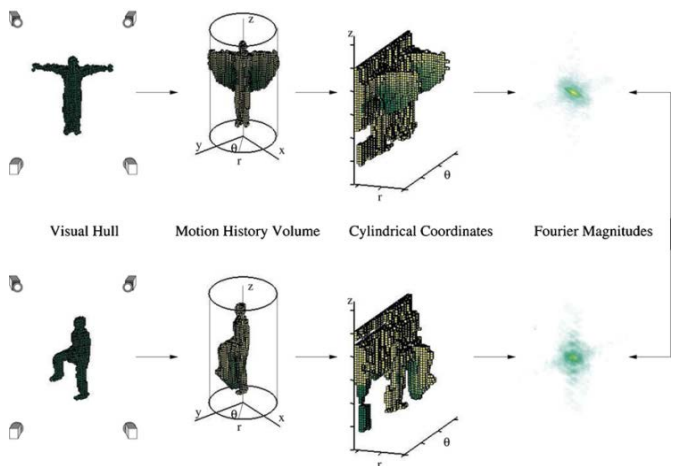

(f)

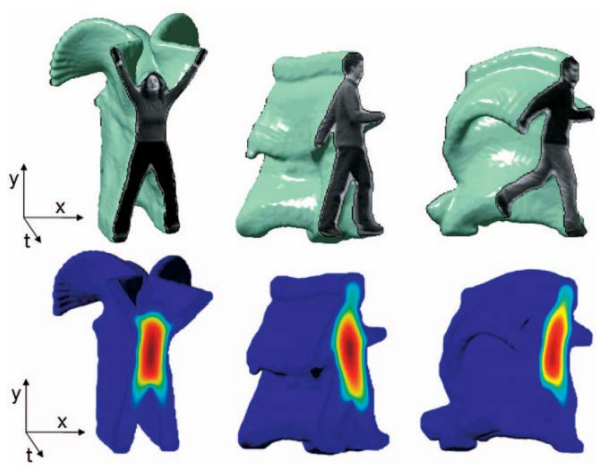

(g)

Fig. 10 Different approaches for action recognition: a) detecting periodic activity using low-level motion features (Polana and Nelson [53] @ 1994 IEEE), b) MEI and MHI (Bobick and Davis [52] @ 2001 IEEE), c) 2D spatiotemporal trajectory (Cen Rao et al. [59] @ 2002 Kluwer Academic Publishers.), d) STVs (Yilmaz and Shah [58] @ 2005 IEEE), e) geometric invariants computed from five points that lie in a plane (Parameswaran and Chellappa [61] @ 2006 Springer), f) MHVs (Weinland et al. [57] @ 2006 Elsevier), g) space-time shapes (Gorelick et al. [60] @ 2007 IEEE)

To conclude up this subsection, it is important to indicate that the different approaches and algorithms used to solve different stages of the problem of human motion understanding may be suitable for different situations, which means that there is no single approach or algorithm to be claimed superior to another one in all situations. Each one of them has its pros and cons. Moreover, many approaches or algorithms may be combined together to yield a better recognition results. For example, if we consider the classification of action recognition algorithms into discriminative and generative algorithms, we find that a generative model learns the joint probability distribution of all variables (i.e., observations and action classes), and thus it learns to model an action class with all of its variations. On the other hand, a discriminative model learns the conditional probability distribution, and thus it learns the differences between action classes in order to discriminate between them. A well-known comparison [33] between generative and discriminative algorithms shows that the discriminative algorithms have lower error rates when dealing with larger training sets, while generative algorithms converge to their higher asymptotic errors much faster. Moreover, generative algorithms have been found to show more flexibility when dealing with small datasets or handling partially labeled data and they are better suited for learning complex patterns, while the discriminative approaches are best suited for classification purposes. If we take HMM as an example of generative algorithms, we find that it combines together the benefits of a temporal evolution model (such as FSM) and a probabilistic model (such as BN) [17]. However, CRF, an example of discriminative algorithms, outperforms HMM for similar action recognition tasks because of their ability to choose an arbitrary dependent abstraction scheme (which may be defined as a categorization of low-level inputs into pixel-based, object-based and logic-based abstractions), and that the abstraction feature selection can consider any combinations of past and future observations [17]. Moreover, many researchers advise to use CRFs instead of HMMs to deal with view-invariant human motion analysis because of their ability of modeling dependencies between features and observations [18]. On the other hand, CRFs take longer time for learning than HMMs do.

**4- Datasets**

Recording human motion datasets with specific constraints and requirements is not an easy task. Rather it is very time-consuming. Comparison of different approaches used in the field of human motion analysis and recognition is very difficult if each approach use its own dataset. Thus, the need for public human motion datasets that covers several aspects of the human action, activities and interactions in different scenarios and situations is very crucial for progressing in this field because these datasets will save time and resources and facilitate comparison of different approaches. To what degree may human observers agree on spatial (e.g., size of an individual, position of a group, etc), temporal (e.g., initial appearance and exit times of a person), and behavioral (e.g., loitering or browsing) observations is also an important factor to consider in generating ground-truth data [36] because there is no single standard technique for doing so. Thus, providing ground-truth data with the recorded datasets is even more desirable but is also more tedious to obtain. It may be very helpful in describing different datasets to include information like:

- Performed actions (e.g., number of actions; what actions are performed; are actions realistic or semi-realistic; interclass variation of the performed actions; are the actions performed in different styles; are the actions performed in multiple views or in a single one; number of performers and variations among them considering age, gender, height and weight; does the performer appear alone in the scene; whether trained performers are involved; whether full-body is involved in the action or just some body parts; is there interaction; is occlusion encountered, etc).
- Environments Constraints (e.g., controlled or uncontrolled, static or dynamic background, indoor or outdoor, lighting conditions, weather conditions, etc).
- Types of cameras involved (e.g., visible, infrared), their number, stationary or moving, gray or color images, is there overlapping between cameras or not, etc.
- Video file characteristics (e.g., video resolution, frame rate, duration, video file extension, file size, raw or compressed data, etc).
- Whether there are any additional data (i.e., ground-truth data such as segmented silhouettes, bounding boxes, list of objects in every frame, automatic or human-made annotations, etc).
- Acknowledgment about the institution that build the dataset, original goal of the dataset, year of creation, website to download, whether downloading requires a license agreement, descriptive paper if any, etc.

Recent surveys about human motion analysis, action, or activity recognition started to have a section in each one of them for discussing some of these public datasets or to make simple comparisons between them [18, 19, 20, 21, 24]. For example, Weinland et al. [20] discussed three datasets, which are KTH, IXMAS and Weizmann, and reported the action recognition rates of about 36 papers that used one or two of these action datasets. They declared that the original papers of the KTH and Weizmann datasets have recognition rates of 71.7% and 99.6%, respectively. They also showed that the highest recognition rates achieved in those 36 papers (published from 2004 to 2009) are: 91.8% (KTH dataset with the same evaluation criteria used in the original paper), 94.2% (KTH dataset with leave-one-out cross-validation), 82.8% (IXMAS dataset with 2D approaches), 98.8% (IXMAS dataset with 3D approaches), and 100% (Weizmann dataset). Another example can be seen in [24] where Holte et al. indicated a number of multi-view datasets but focused only on two datasets: IXMAS and i3DPost. They reported the action recognition rates of 18 papers (published from 2006 to 2011) that used either 2D approaches or 3D approaches evaluated on the IXMAS dataset indicating that 3D approaches are the top performing methods, especially those that are based on motion history volumes (MHVs) with highest accuracy of 98.8%, and that the best performing 2D methods is based upon motion history image features and ballistic dynamics with accuracy of 87%. The authors also reported action recognition rates of 3 papers (published in 2009,2010,2011) evaluated on the i3DPost dataset indicating again that the approach based on 3D motion information (92.19%) outperforms the other two 2D-approaches (90.88%, 90%). And thus, they concluded that using full reconstructed 3D information is superior to using 2D image data captured from multiple views but that comes on the cost of more computations. Guerra-Filho and Biswas [37] introduced a short survey about eight MOCAP "motion capture" datasets declaring original goal, number of performers, number of actions, other modalities if used such as audio, advantages and disadvantages. They also described the construction of their human motion database using a systematic controlled capture methodology, known as cognitive and parametric sampling, to obtain the structure necessary for the quantitative evaluation of several motion-based research problems. Their database is organized into five datasets: the praxicon dataset, the cross-validation dataset, the generalization dataset, the compositionality dataset, and the interaction dataset. Chaquet et al. [35] reported a total of 68 datasets that are available for human action and activity recognition (an interested reader should find this survey very helpful in choosing what dataset to use in his/her research). They categorized these datasets into three categories. The first one is assigned to heterogeneous actions, which include 28 of these datasets. They described these datasets in more details than the others, where they included for each one of them, the following information: dataset identification (e.g., the institution that built the dataset, the year of creation, the website to download, the descriptive paper if any), original goal, application domains, ground truth data (e.g., segmented silhouettes, bounding boxes, foreground masks, 3d models/volumes, action annotations, etc), context (e.g., the performed set of actions or activities, the number of actions, the interactions types if exist "person-person or person-object", the type of sceneries "indoors or outdoors", the number of actors, the number of views, camera movement, etc), example video frames, and finally a list of papers that used the dataset. For the rest of the datasets which they categorized into specific actions (e.g., detection of abandoned objects, crowd behavior, pose or gesture recognition, gait analysis, falls detection, activity of daily living "ADL", etc), or others (e.g., motion capture, infrared, etc.), they included the following information: the year of creation, the website to download, the descriptive paper if any, and the category of the dataset. The authors declare that the oldest dataset, described in their survey, has been recorded in 2001 and that 86% of these datasets were created from 2005 onwards. They also indicated that there has been an increase in recording more datasets in recent years, e.g., 9 of the 28 heterogeneous-actions datasets have been recorded in the time window from 2010 to 2012.

We list here some of the datasets that are suitable for human motion analysis as follows: **KTH** "*Kungliga Tekniska högskolan*" dataset; Weizmann datasets; **CAVIAR** "*Context Aware Vision using Image-based Active Recognition*" dataset; **ETISEO** "*Evaluation du Traitement et de l'Interpretation de Sequences Video*" dataset; **PETS** "*Performance Evaluation of Tracking and Surveillance*" datasets; **ALOV++** "*Amsterdam Library of Ordinary Videos*" dataset; **IXMAS** "*INRIA Xmas Motion Acquisition Sequences*" dataset; **UCF** "*University of Central Florida*" Human Actions datasets (e.g., UCF Sports Action, UCF YouTube Action, UCF101, UCF50, UCF-iPhone, UCF Aerial Action, UCF-ARG); Hollywood dataset; Hollywood2 dataset; **HMDB51** "*Human Motion DataBase*"; VideoWeb dataset; Olympic Sports dataset; TV Human Interaction dataset; i3DPost Multi-view dataset; BEHAVE dataset; **MSR** "*Microsoft Research*" Action Recognition datasets; **ViSOR** "*Video Surveillance On-line Repository for Annotation Retrieval*" dataset; **VIRAT** "*Video and Imagery Retrieval and Analysis Toolkit*" dataset; **MuHAVi** "*Multicamera Human Action Video Data*" dataset; **CASIA** "*Chinese Academy of Sciences, Institute of Automation*" Action Recognition Database; **CASIA** Gait Recognition Database (which include four datasets A, B, C, and D); **UIUC** "*University of Illinois at Urbana-Champaign*" Action dataset; **CMU** "*Carnegie Mellon University*" datasets (e.g., CMU MMAC, CMU MoBo, CMU Graphics Lab Motion Capture); **UT** "*University of Texas*" datasets (UT-Interaction, UT-Tower); **URADL** "*University of Rochester Activities of Daily Living*" dataset; Korea University Gesture dataset; Cambridge Hand Gesture dataset; Biological Motion library dataset; Buffy pose classes package dataset; Buffy Stickmen dataset; **CANDELA** "*Content Analysis and Network DELivery Architectures*" dataset; Drinking and Smoking dataset; Gait Based Human ID Challenge Problem dataset; HDM05 dataset; **HMD** "*Human Motion Database*"; HumanEVA datasets; Human Identification at a Distance dataset; **ICS** "*Intelligent Cooperative Systems*" Action Database; **IEMOCAP** "*Interactive Emotional Dyadic Motion Capture*" dataset; **i-LIDS** "*Imagery Library for Intelligent Detection Systems*" datasets; Keck Gesture dataset; LDB dataset; WAR dataset; **MPII** Cooking Activities dataset; Multiple cameras fall dataset; **OTCBVS** "*Object Tracking and Classification in and Beyond the Visible Spectrum*" Benchmark dataset; **POETICON** dataset (*an European project that explores the "poetics of everyday life", i.e. the synthesis of sensorimotor representations and natural language in everyday human interaction for developing a computational mechanism for generalization and generation of new behaviors in robots*"; **NATOPS** "*Naval Air Training and Operating Procedures Standardization*" dataset; **RVL-SLLL ASL** "*Robot Vision Lab and Sign Language Linguistics Lab for American Sign Language Database*"; TUM Kitchen dataset; **ViHASi** "*Virtual Human Action Silhouette*" dataset; Collective Activity Dataset; **UMPM** "*Utrecht Multi-Person Motion*" dataset; WorkoutSU-10 Exercise dataset; **VSD** "*Violent Scene Dataset*"; **CLEAR** "*CLassification of Events, Activities, and Relationships*" dataset; **SPEVI** "*Surveillance Performance EValuation Initiative*" datasets; **TRECVID** "*Text REtrieval Conference Video Retrieval Evaluation*" datasets; "mocapdata.com" datasets.

Many of the previous datasets were originally designed for human action, interaction or activity recognition (we will refer to these datasets as action datasets as they are generally known), others are designed for other purposes such as tracking, pose estimation, person identification, action retrieval, crowd behavior analysis, detecting abandoned objects, detecting falls, etc, but they contain sequences that are also suitable for action recognition. We will give a quick look at a sample from the above action datasets that contains semi-realistic action recognition datasets "simple datasets", more realistic and challenging action recognition datasets, interaction datasets and multi-view datasets.

- **Simple Action Datasets:** A good representative of this category is the Weizmann dataset, which is considered the easiest dataset mentioned here. Another example is the KTH dataset, which add more challenges such as scale variation. KTH dataset is recorded in 2004 by the Swedish Royal Institute of Technology "*Kungliga Tekniska högskolan*". The dataset contains six actions which are walking, jogging, running, boxing, hand waving, and hand Clapping. These actions were performed by 25 subjects in four different scenarios: outdoors, outdoors with scale variation, outdoors with different clothes, and indoors. All sequences were taken over homogeneous backgrounds with a static camera with *25*fps frame rate. The sequences were downsampled to *160x120* resolution and have a length of four seconds in average. They can be downloaded from [143] where they have been stored in the AVI format using DIVX compression, but uncompressed version is also available on demand.

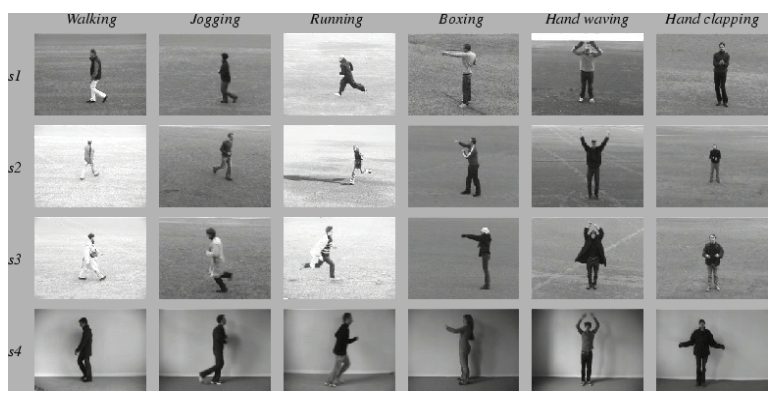

Fig. 11 Example frames from the KTH dataset

- **More Realistic and Challenging Action Datasets:** E.g., CAVIAR, ETISEO, Hollywood datasets UCF Human Actions datasets, HMDB51, MSR Action, Olympics Sports, ViSOR, VIRAT, etc. For example, the Hollywood 2 dataset, created in 2009, contains 12 classes of human actions and 10 classes of scenes distributed over 3669 video clips and approximately 20.1 hours of video in total. The 12 human actions are: answer phone, driving car, eat, fight, get out car, hug person, hand shake, kiss, run, sit down, sit up, and stand up. The dataset's video clips are extracted from 69 movies. It contains about 150 samples per action class and 130 samples per scene class in training and test subsets. Download is available from [144].

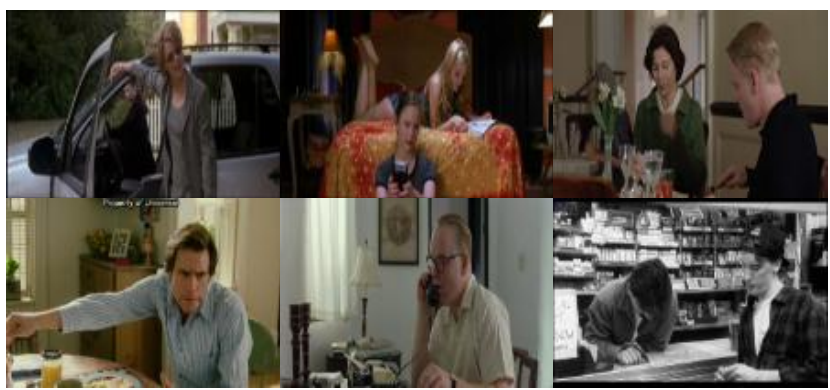

Fig. 12 Example frames from the Hollywood 2 dataset

The Olympics Sports dataset, created in 2010, contains videos of athletes practicing different sports. The authors obtained the video sequences from YouTube and annotated their classes using Amazon Mechanical Turk. The dataset contains 16 sports which are: high jump, long jump, triple jump, discus throw, hammer throw, javelin throw, shot put, platform diving, springboard diving, bowling, basketball lay-up, tennis serve, snatch (weightlifting), clean and Jerk (weightlifting), pole vault, gymnastic vault. All video sequences are stored in the SEQ video format. Download is available from [145].

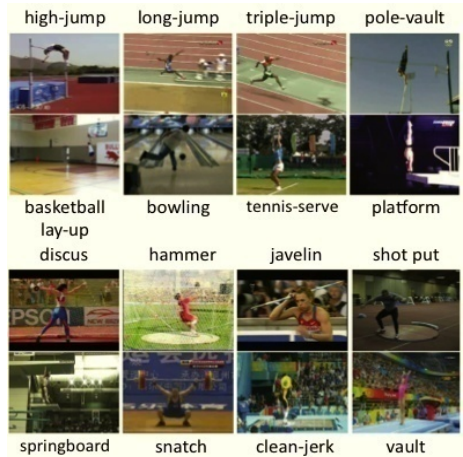

Fig. 13 Example frames from the Olympics Sports dataset

The HMDB51 dataset, created in 2011, is collected from various sources, mostly from movies, and a small proportion from public databases such as the Prelinger archive, YouTube, and Google videos. The dataset contains 6849 clips that are divided 51 action classes in the dataset, each containing a minimum of 101 clips. The actions classes are grouped in five categories: general facial actions (e.g., smile, laugh, chew, talk), facial actions with object manipulation (e.g., smoke, eat, drink), general body movements (e.g., cartwheel, clap hands, climb, climb stairs, dive, fall on the floor, backhand flip, handstand, jump, pull up, push up, run, sit down, sit up, somersault, stand up, turn, walk, wave), body movements with object interaction (e.g., brush hair, catch, draw sword, dribble, golf, hit something, kick ball, pick, pour, push something, ride bike, ride horse, shoot ball, shoot bow, shoot gun, swing baseball bat, sword exercise, throw), body movements for human interaction: fencing, hug, kick someone, kiss, punch, shake hands, sword fight). In addition to labeling all action classes, each clip is annotated with an action label as well as a meta-label describing the property of the clip such as visible body parts, camera motion, camera viewpoint, number of people involved in the action, video quality. Download is available from [146]. Refer to [39] for more information about the dataset.

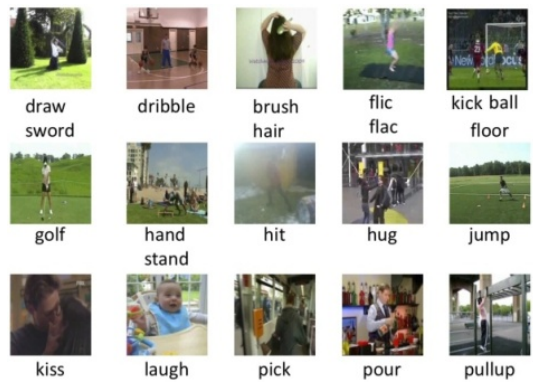

Fig. 14 Example frames from the HMDB51 dataset

The UCF101 action recognition dataset, collected from YouTube, is an extension of the UCF50 dataset. It has 13320 video sequences covering 101 action classes. The videos in each action class are grouped into 25 groups, where each group can consist of 4-7 videos. The videos from the same group may share some common features, such as similar background, similar viewpoint, etc. The action classes can be divided into five categories: body-motion only, human-object interaction, human-human interaction, playing musical instruments, and sports. The UCF101 dataset show large diversity in terms of actions, cluttered background, illumination conditions, camera motion, viewpoint, object scale, object appearance and pose, etc.

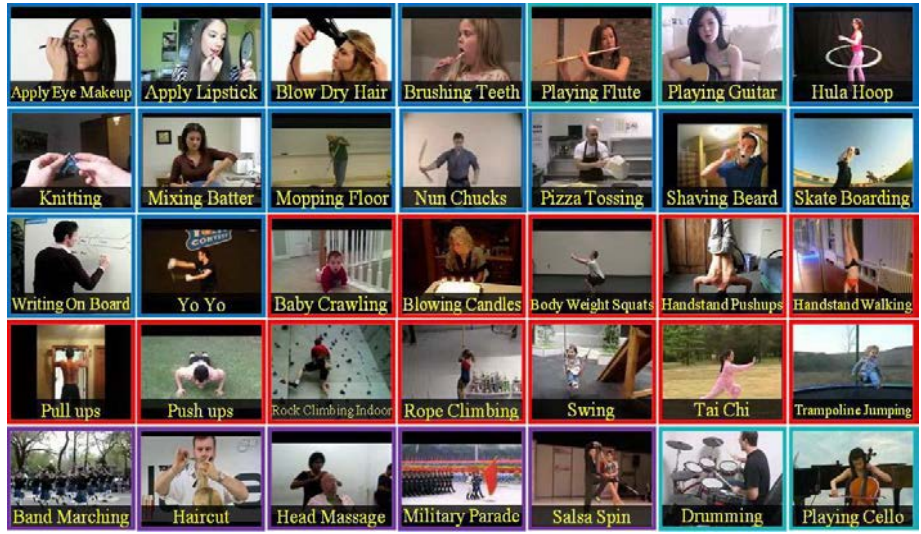

Fig. 15 Example frames from the UCF101 action dataset

The action classes for the UCF101 dataset are: apply eye makeup, apply lipstick, archery, baby crawling, balance beam, band marching, baseball pitch, basketball shooting, basketball dunk, bench press, biking, billiards shot, blow dry hair, blowing candles, body weight squats, bowling, boxing punching bag, boxing speed bag, breaststroke, brushing teeth, clean and jerk, cliff diving, cricket bowling, cricket shot, cutting in kitchen, diving, drumming, fencing, field hockey penalty, floor gymnastics, frisbee catch, front crawl, golf swing, haircut, hammer throw, hammering, handstand pushups, handstand walking, head massage, high jump, horse race, horse riding, hula hoop, ice dancing, javelin throw, juggling balls, jump rope, jumping jack, kayaking, knitting, long jump, lunges, military parade, mixing batter, mopping floor, nun chucks, parallel bars, pizza tossing, playing guitar, playing piano, playing tabla, playing violin, playing cello, playing daf, playing dhol, playing flute, playing sitar, pole vault, pommel horse, pull ups, punch, pushups, rafting, rock climbing indoor, rope climbing, rowing, salsa spins, shaving beard, shotput, skate boarding, skiing, skijet, sky diving, soccer juggling, soccer penalty, still rings, sumo wrestling, surfing, swing, table tennis shot, tai chi, tennis swing, throw discus, trampoline jumping, typing, uneven bars, volleyball spiking, walking with a dog, wall pushups, writing on board, and yo yo. Download is available from [147]. Refer to [38] for more information about the dataset.

- **Interaction Datasets:** E.g., TV Human Interactions, UT-Interaction, BEHAVE, etc. For example, TV Human Interactions dataset consists of 300 video clips collected from over 20 different TV shows and containing 4 interactions: handshakes, high fives, hugs and kisses, as well as clips that don't contain any of the interactions. The authors of this dataset used frame-based annotations in all videos of this dataset. They annotated the following: the upper body of people (with a bounding box), the discrete head orientation (profile-left, profile-right, frontal-left, frontal-right and backwards), and the Interaction of each person. This dataset was recorded in 2010 by the visual geometry group of engineering science department, university of Oxford. Download is available from [148].

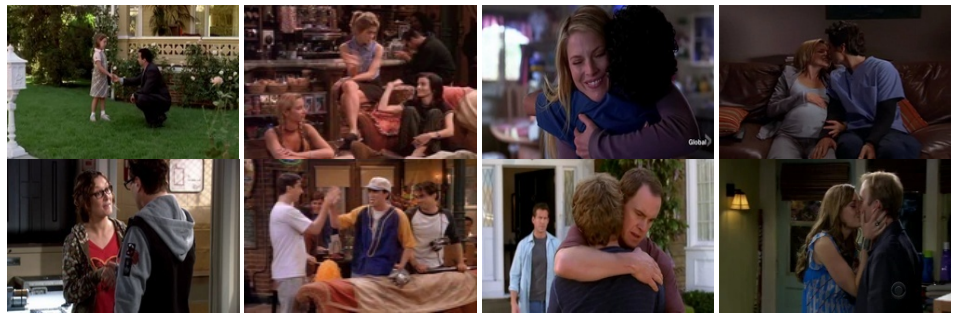

Fig. 16 Example frames from the TV Human Interactions dataset (handshakes, high fives, hugs and kisses)

**Multi-view Datasets:** E.g., IXMAS, i3DPost, MuHAVi, VideoWeb, HumanEva-I, CASIA Action Database, CASIA Gait Recognition Dataset B, CMU Motion database, UCF-ARG, etc. For Example, i3DPost dataset is a multi-view/3D human action/interaction dataset created in 2009 by the university of Surrey and CERTH-ITI (Centre of Research and Technology Hellas Informatics and Telematics Institute) within the i3DPost project. It contains 12 motions plus 6 basic facial expressions performed each by 8 subjects. These motions are walk, run, jump forward, jump in place, bend, one hand wave, sit down-stand up, walk-sit down, run-fall, run-jump-walk, two persons handshaking, and one person pulls another. The uncompressed videos have been recorded using 8 synchronized and calibrated cameras with *1920x1080* resolution at 25Hz. A 3D mesh model is provided for each video frame describing the respective 3D human body surface. Refer to [40] for more information about the dataset. Download is available from [149] but after a license agreement.

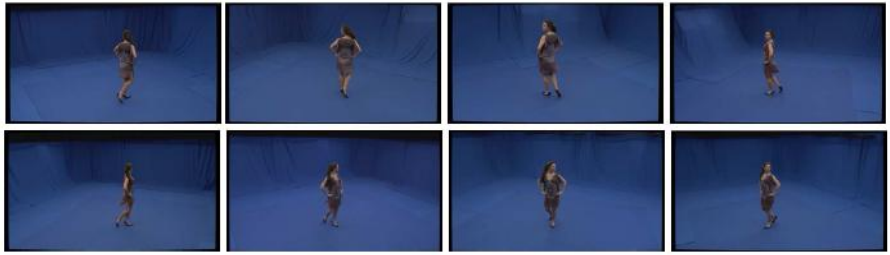

Fig. 17 i3DPost multi-view dataset (eight different views of the same instant)

Another example is the CASIA action database created in 2007, which is a collection of sequences of human activities captured by video cameras outdoors from different views. There are 8 classes of single-person actions (walk, run, bend, jump, crouch, faint, wander and punching a car) that are performed each by 24 subjects. There are also 7 classes of two person interactions (rob, fight, follow, follow and gather, meet and part, meet and gather, overtake) that are performed by every 2 subjects. The total number of sequences is 1446 with varying durations from 5 seconds to 30 seconds. All video sequences were taken simultaneously with static three non-calibrated cameras from different views (horizontal view, angle view, and top down view). The sequences were compressed with the Huffyuv code in AVI format at 25fps with a resolution of *320x240*. Download is available from [150] after assigning an agreement.

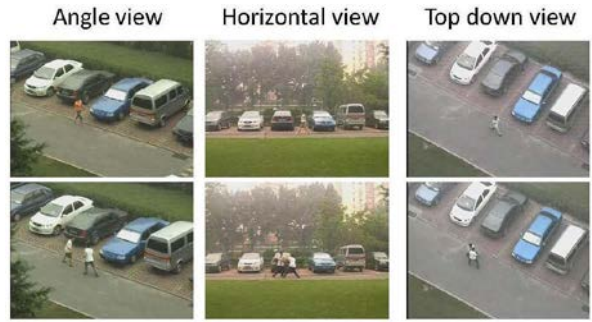

Fig. 18 CASIA action database (the first raw depicts a walking person and the second row depicts two person fighting each other)

Another fast look can be given to the international series of PETS workshops that motivate many researchers in the community of computer vision to use their standard datasets and ground-truth data in order to facilitate comparison and evaluation of different algorithms in the domain of surveillance and tracking systems. While the main goal of the PETS datasets is people tracking, counting, and surveillance, not action recognition, but their sequences contain some actions such as walking, running, loitering, and attended luggage removal "theft" that are also suitable for action recognition. Also, recent PETS datasets focus more on crowd behavior analysis and event detection. Previous datasets can be downloaded from [151], and recent ones (PETS 2013 and PETS 2009 "both are the same") that contain different crowd activities can be downloaded from [152].

It is clear that early datasets tend to deal with semi-realistic actions performed with simple and static backgrounds such as Weizmann and KTH datasets. These datasets cover few actions, which are also performed by professionals. They also provide simple ground-truth data. On the other hand, recent datasets deal with more realistic actions, interactions and activities in more complex situations and conditions, e.g., Hollywood datasets, UCF101, HMD51, VIRAT, TV human Interaction, etc. Actions performed by directed performers were also minimized and the majority of actions were performed by the general population. Many of these datasets consider larger number of actions and various interactions in diverse scenes, encountering frequent incidental movers and background activities. Recent datasets also provide high quality ground-truth data due to the population of xml-based description languages such as ViPER-GT "Video Performance Evaluation Resource Ground Truth", which simplify the visual annotation process, e.g., rectangles denoting locations of people on screen, resulting in richer descriptions of each frame in the video sequence. In the end, we can conclude that the progress in the field of human motion analysis and understanding can be mirrored by the maturity and complexity levels of the datasets being used.

## 5- Discussion

The research in computer vision has started since 1960s, and its interest in recognizing and understanding human motion has emerged in the early 1980s or late 1970s. Since then, computer vision researchers, in their quest to solve this problem, have started to use various techniques and algorithms found in many disciplines, with which it shares its boundaries, e.g., artificial intelligence, image processing, signal processing, applied mathematics, statistics, geometry, neural networks, physics, neuroscience, psychophysics, biological vision, etc. However, they have faced several problems in detecting, segmenting, analyzing and recognizing different human motions from video sequences. Some of these problems are: data loss resulted from projection of a 3D scene to a 2D image, image noise (e.g., salt and pepper noise, fixed pattern noise, banding noise), dynamic or cluttered background, lighting conditions (e.g., indoor, outdoor, morning, night, etc), weather conditions (e.g., rainy, foggy, sunny, windy, etc), illumination change, light reflections, shadows, temporal textures motion (e.g., tree leaves motion, waving clothes in the air, flying birds, etc), object appearance change, object shape change, wearing excessively sloppy clothes, full or partial occlusion (whether it is an object-to-object occlusion or a scene-to-object occlusion), number of humans in the scene, articulated type of the human motion, complex motion, camera motion, distance of the object from the camera and if zooming is required or not, different viewpoints of the performed motion, data fusion from multiple cameras or different types of sensors. For actions and activities recognition, the researchers encounter extra problems such as the interclass variations between different performers (e.g., speed, pace, anthropometric variation), different execution rates of the same action by the same performer, performing the same action with slight variations (e.g., walking while carrying objects, walking on crutches or with a walking stick, walking while holding on to a walker, nordic walking, walking with a dog, etc), distinguishing between similar classes that are close in performance (e.g., walking, line walking, marching [51]), performance similarity of different classes.

In order to overcome many of these difficulties, researchers imposed some constraints on background appearance, and/or object appearance, and/or object motion, and/or number of objects, etc. The suitability of a certain algorithm depends on the proposed constraints. Some of these constraints are: using controlled environment (e.g., indoors, markers placed on the performer), constant lighting, static background, uniform or simple background, tight clothes, disagreement between the colors of the background and the object clothes, no occlusion, no camera motion, object motion is parallel to the motion plane, moving on a flat ground plane, motion periodicity, constant speed, manual initialization of the human body pose, reducing dimensions of human body modeling, one person in the scene, known camera parameters, etc. Another way to deal with these difficulties is to introduce more information about the scene such as using multiple cameras, audio sensors, radar, ultrasonic, etc. Although, for example, using multiple cameras proves effectiveness in handling the occlusion problem, building a 3D model of the object, finding more suitable features, etc, but this, of course, adds more costs, computations, and complexities to the problem such as increasing installation costs of using more cameras, increasing processing time, calibrating cameras, searching for features in each camera image separately and then combining the information or combining the information early in order to reconstruct a 3D model, or selecting an active viewpoint to determine which camera is the most suitable for more clearer information. Moreover, fusing different types of features (e.g., color, shape, position) from different viewpoints of an object is also not an easy task. Fortunately, recent technological advances in real-time image capture, transfer, and processing have been an encouraging factor to lessen the burden of these computations and costs, and also further the research on human motion analysis through using more powerful and complex algorithms.

In the field of human motion recognition and understanding, behavior recognition may be the hardest problem to implement because the same behavior may have several different interpretations depending on many factors such as scene, task, object, and cultural contexts, attention, intention, etc.

The selection of the most discriminative features of an object is a very important issue in designing algorithms. For example, the features should not change significantly over time, be robust to transformations (e.g., translation, rotations, scaling), be robust to illumination change conditions (e.g., edges and textures are less sensitive to illumination change compared to color feature), etc. Combining multiple and complementary features is also of great importance to ensure the continuity of the algorithm. It is also better to have an algorithm that can select the appropriate features online automatically.

In the same vein, incorporating prior and contextual information is very helpful to adjust the algorithm to a particular scenario in which it is used. It is intuitive that observing scenes provide us not only with the information that our eyes capture but also with any knowledge that we may have earlier about the observed scenes. This helps us in many directions such as quickly guiding our attention to regions of interest, interpreting and understanding the scenes correctly, etc. In other words, we can say that context provide us with inference information that can help our understanding about the scene. Therefore, computer vision systems should also be able to use contextual information. Computer vision lacks a precise and common definition of the term "context"; researchers understand the term tacitly; however, we will use the definition of Strat [41] that defines it as "any and all information that may influence the way a scene is perceived". Classification of contextual information differs according to the domain of application and the goal of the computer vision system [41]. The usage of context in the field of human motion recognition and understanding is still small and mostly limited to scene and object contexts [42]. A recent study by Reddy and Shah [44] demonstrates that scene context is a very important feature to perform action recognition on very large datasets. Moreover, it shows that the proposed scene context descriptor becomes more discriminative and that the motion descriptors become less discriminative as the number of action categories increases.

The role of attention to decrease the search space or to optimize the search methodology is very vital to make computer-vision systems able to achieve the goal of human-like performance. Attention may be defined as "a set of strategies that attempts to reduce the computational cost of the search process inherent in visual perception" [43]. Tsotsos [43] indicated that he and others have shown theoretically that without attention, particularly without the use of task-directed attention to guide processing, the vision problem in the general case would be exponential complex. In literature, many assumptions or constraints imposed by researchers to ease the problem of motion recognition and understanding have led to reduce or eliminate the role of attention (e.g., fixed camera system, pre-segmentation, clean background, knowledge of the task domain), but different kinds of predictions (e.g., object location predictions) can also be seen as strategies that help to reduce the search space and hence fall within the proposed definition of attention. However, attention can play a more prominent and effective role in solving the general case of the vision problem. For example, designing multi-purpose systems, by combining together a number of single-purpose systems and using a smart switching technique to apply the most suitable system at the current situation, would not be scaled well; instead, optimizing the basic processing methods for the task at hand would be more fruitful and also simulating what the human visual system seems to do (many biological evidences indicate the effects of the attentive processing on all aspects of visual perception including motion) [43].

The role of intention in understanding human actions and behaviors is also very significant. It has attracted many researchers in different disciplines to study it. For example, In order to know how people understand intentional action, some evolutionary theorists such as Tomasello et al. [125] have proposed a model of what intentional action is. The proposed model is based on the principles of control systems, where goal, action, and perceptual monitoring are the basic components of a system that serves to regulate human's behavioral interactions with the environment. In the proposed model, an intention is defined as a plan of action that the person chooses to achieve his goal, and thus, an intention includes both a means (action plan) as well as a goal. On other sides, we find controversial debates between neuroscientists and philosophers about conscious intention or conscious free will [140, 141]. Philosophers suggest that our conscious intentions cause our actions and that implies both mind-body causation and dualism (i.e. the intention would drive the motor areas of the brain, and thus, the body muscles to realize the action). In other words, they suggest that conscious experience causes brain activity not vice versa. On the other hand, neuroscientists claim that the subjective experience of conscious intention is no more than an illusion, and that our actions are initiated by unconscious mental processes that take place in the frontal and parietal motor areas earlier than we become aware of our intention to act. Haggard [140] defined the term "intention" as a collection of several distinct processes within the chain of information processing that translates desires and goals into behavior. An interesting fMRI experiment by Soon et al. [142] revealed that an intention can be encoded in brain activity of prefrontal and parietal cortex up to 10s before it enters awareness, and that this delay indicates the presences of a high-level control network that begins to prepare an upcoming intention long before it enters awareness. To conclude this point, we want to indicate that the role of intention should not be neglected when designing computer vision systems that are capable of understanding different human activities and behaviors.

In designing algorithms to solve the problem of human motion recognition and understanding, several factors should be considered such as: the generality of the proposed algorithm or its applicability to a specific domain or context, efficiency or real time processing requirements (e.g., it is very critical to some applications such as intelligent driver assistance systems, and it is required in applications such as assisted living, human-computer interaction, gesture-based interactive games, smart environments, visual surveillance, but it is not essential in applications like entertainment industry or sports motion analysis), robustness (which is important for continuity and can be tested through using a large amount of data, using different performers, employing dynamic environments, changing conditions, etc), accuracy or precision (high accuracy is required in sports motion analysis, movies industry, etc, but for applications such as human-computer interaction, gesture-based interactive games it may vary between medium and high).

We have mentioned earlier that using common public datasets with their ground-truth data is very helpful in comparing different algorithms, but there are also other factors that can make the comparison more useful and profound for progressing in the field such as using a common evaluation criteria (e.g., speed, performance robustness and accuracy), and a common evaluation methodology or experimental setup (e.g., using the dataset split as that of the original paper [the dataset may be divided into two or three disjoint sets: training/testing sets or training/validation/testing sets, respectively], or using the same validation technique such as leave-one-out cross-validation). Other important factors for comparing different approaches are to tell whether the whole dataset is used or just a subset of it, and whether a part of the video is used or the whole video.

In literature, different algorithms have been developed for dealing with specific-domain applications, but the ultimate goal of computer vision researchers in the field of human motion analysis and recognition is to enable computer systems to have human-level recognition of any types of motion. So, calls for a unified framework that benefits all motion recognition tasks are growing stronger. For example, Aggarwal and Ryoo [42] suggest building such a unified framework. They believed that the main challenges for building this unified framework are: 1- constructing a set of features from which a suitable set of these features can be automatically chosen when the system handles a certain problem, the features must describe the system sufficiently, and thus, a careful extraction of the features is very important (e.g., features that handles multiple viewpoints, features from low resolution videos, features that are immune to noise and variations such as high-dimensional local spatio-temporal features, etc), 2- considering contexts, such as, social, cultural, and biological contexts, beside using scene and object contexts, 3- understanding human intention in order to understand their behavior (e.g., differentiating between real fight and martial arts sparring requires knowing the intentions of the persons engaged in the action), and that requires a formal representation of the concept to be constructed, and then model its relationship with human activities mathematically.

## 6- Conclusion

To make computer vision systems able to understand human activities and behaviors, the research in computer vision has changed its direction a number of times. First, the researchers understand that they have to deal with low level processing before attempting to solve high-level problems (e.g., the MIT copy-demo). Then, the influential work of Marr [135], as was previously stated, has emphasized the role of the human visual system in analyzing and understanding human motion (i.e., the human visual system is able to recover the 3D structure of objects from their changing 2D images formed on the retina). This leads many computer vision researchers to recover 3D structures of objects in order to solve vision problems concerning with human motion representation and recognition. Thus, the popular approach was then "shape from X" that operates on recovering the lost dimension resulted from projection of the 3D scene on to the 2D image (X here may refer to motion, or texture, or shading, or stereo) [45, 46]. Then, another approach, known as "motion-based recognition", has proved its strength and effectiveness since the mid 1990s [1]. Before the year 2001, the latest computer vision systems had used more advanced techniques that are based on probabilistic models but issues like automatic initialization, robustness and recovering from failure were not addressed carefully [5]. In the early 21st century, most approaches were interested in simple action recognition [11]. In the last decade, the trend in the field has been escalating from dealing with semi-realistic actions performed with simple and static backgrounds to deal with more realistic actions, interactions and activities in more complex situations and conditions. For example, there are many algorithms that can deal with multiple persons in the scene, occlusion, view invariance, execution rate invariance, anthropometric invariance, etc. Moreover, almost all public datasets have been created in this last decade. To sum up the current progress in the field, we can say that the lower levels of data processing such as object detection and tracking have reached a reasonable degree of maturity, but for higher levels such as activity or behavior recognition, there is still much more to be done. For more progress in the field, computer vision researchers may benefit from combining different techniques together such as combining generative and discriminative approaches. They can also benefit from adopting techniques such as online learning which allow adaptability to the changing conditions of a problem. Finally, we want to indicate that both the calls for multi-disciplinary projects and the calls for a unified framework that benefits all motion recognition tasks would also be of great help to achieve more advances in the field.

**References**
[1] Cedras, Claudette, and Mubarak Shah. "Motion-based recognition a survey." *Image and Vision Computing* 13.2 (1995): 129-155.
[2] Aggarwal, J. K., Cai, Q., Liao, W., & Sabata, B. "Nonrigid motion analysis: Articulated and elastic motion." *Computer Vision and Image Understanding* 70.2 (1998): 142-156.
[3] Gavrila, Dariu M. "The visual analysis of human movement: A survey." *Computer vision and image understanding* 73.1 (1999): 82-98.
[4] Aggarwal, Jake K., and Qin Cai. "Human motion analysis: A review." *Computer Vision and Image Understanding,* Vol. 73.3, March, pp. 428–440, 1999.
[5] Moeslund, Thomas B., and Erik Granum. "A survey of computer vision-based human motion capture." *Computer Vision and Image Understanding* 81.3 (2001): 231-268.
[6] Wang, Liang, Weiming Hu, and Tieniu Tan. "Recent developments in human motion analysis." *Pattern recognition* 36.3 (2003): 585-601.
[7] Buxton, Hilary. "Learning and understanding dynamic scene activity: a review." *Image and vision computing* 21.1 (2003): 125-136.
[8] Wang, Jessica JunLin, and Sameer Singh. "Video analysis of human dynamics—a survey." *Real-time imaging* 9.5 (2003): 321-346.
[9] Aggarwal, J. K., and Sangho Park. "Human motion: Modeling and recognition of actions and interactions." *3D Data Processing, Visualization and Transmission, 2004. 3DPVT 2004. Proceedings. 2nd International Symposium on*. IEEE, 2004.
[10] Hu, Weiming, Tieniu Tan, Liang Wang, and Steve Maybank. "A survey on visual surveillance of object motion and behaviors." *Systems, Man, and Cybernetics, Part C: Applications and Reviews, IEEE Transactions on* 34.3 (2004): 334-352.
[11] Moeslund, Thomas B., Adrian Hilton, and Volker Krüger. "A survey of advances in vision-based human motion capture and analysis." *Computer vision and image understanding* 104.2 (2006): 90-126.
[12] Yilmaz, Alper, Omar Javed, and Mubarak Shah. "Object tracking: A survey." *Acm Computing Surveys (CSUR)* 38.4 (2006): 13.
[13] Poppe, Ronald. "Vision-based human motion analysis: An overview." *Computer vision and image understanding* 108.1 (2007): 4-18.
[14] Krüger, V., Kragic, D., Ude, A., & Geib, C. "The meaning of action: a review on action recognition and mapping." *Advanced Robotics* 21.13 (2007): 1473-1501.
[15] Pantic, Maja, Alex Pentland, Anton Nijholt, and Thomas S. Huang. "Human computing and machine understanding of human behavior: A survey." In *Artifical Intelligence for Human Computing*, pp. 47-71. Springer Berlin Heidelberg, 2007.
[16] Turaga, Pavan, Rama Chellappa, Venkatramana S. Subrahmanian, and Octavian Udrea. "Machine recognition of human activities: A survey." *Circuits and Systems for Video Technology, IEEE Transactions on* 18.11 (2008): 1473-1488.
[17] Lavee, Gal, Ehud Rivlin, and Michael Rudzsky. "Understanding video events: a survey of methods for automatic interpretation of semantic occurrences in video." *Systems, Man, and Cybernetics, Part C: Applications and Reviews, IEEE Transactions on* 39.5 (2009): 489-504.
[18] Ji, Xiaofei, and Honghai Liu. "Advances in view-invariant human motion analysis: a review." *Systems, Man, and Cybernetics, Part C: Applications and Reviews, IEEE Transactions on* 40.1 (2010): 13-24.
[19] Poppe, Ronald. "A survey on vision-based human action recognition." *Image and vision computing* 28.6 (2010): 976-990.

[20] Weinland, Daniel, Remi Ronfard, and Edmond Boyer. "A survey of vision-based methods for action representation, segmentation and recognition." *Computer Vision and Image Understanding* 115.2 (2011): 224-241.
[21] Aggarwal, J. K., and Michael S. Ryoo. "Human activity analysis: A review." *ACM Computing Surveys (CSUR)* 43.3 (2011): 16.
[22] Chen, Liming, and Ismail Khalil. "Activity recognition: approaches, practices and trends." *Activity Recognition in Pervasive Intelligent Environments*. Atlantis Press, (2011). 1-31.
[23] Yang, Hanxuan, Shao, L., Zheng, F., Wang, L., & Song, Z. "Recent advances and trends in visual tracking: A review." *Neurocomputing* 74.18 (2011): 3823-3831.
[24] Holte, Michael B., Cuong Tran, Mohan M. Trivedi, and Thomas B. Moeslund. "Human Pose Estimation and Activity Recognition From Multi-View Videos: Comparative Explorations of Recent Developments." *Selected Topics in Signal Processing, IEEE Journal of* 6.5 (2012): 538-552.
[25] Cristani, Marco, Raghavendra, R., Del Bue, A., & Murino, V. "Human behavior analysis in video surveillance: a social signal processing perspective." *Neurocomputing* (2012).
[26] Smeulders, A.; Chu, D.; Cucchiara, R.; Calderara, S.; Dehghan, A.; Shah, M., "Visual Tracking: An Experimental Survey," Pattern Analysis and Machine Intelligence, IEEE Transactions on , vol.PP, no.99, pp.1,1, doi: 10.1109/TPAMI.2013.230
[27] Toyama, Kentaro, John Krumm, Barry Brumitt, and Brian Meyers. "Wallflower: Principles and practice of background maintenance." In Computer Vision, 1999. The Proceedings of the Seventh IEEE International Conference on, vol. 1, pp. 255-261. IEEE, 1999.
[28] Piccardi, Massimo. "Background subtraction techniques: a review." *Systems, man and cybernetics, 2004 IEEE international conference on*. Vol. 4. IEEE, 2004.
[29] Hall, Daniela, Jacinto Nascimento, P. Ribeiro, E. Andrade, P. Moreno, S. Pesnel, T. List et al. "Comparison of target detection algorithms using adaptive background models." In *Visual Surveillance and Performance Evaluation of Tracking and Surveillance, 2005. 2nd Joint IEEE International Workshop on*, pp. 113-120. IEEE, 2005.
[30] Wren, Christopher Richard, Ali Azarbayejani, Trevor Darrell, and Alex Paul Pentland. "Pfinder: Real-time tracking of the human body." Pattern Analysis and Machine Intelligence, IEEE Transactions on 19, no. 7 (1997): 780-785.
[31] Stauffer, Chris, and W. Eric L. Grimson. "Adaptive background mixture models for real-time tracking." Computer Vision and Pattern Recognition, 1999. IEEE Computer Society Conference on.. Vol. 2. IEEE, 1999.
[32] Haritaoglu, Ismail, David Harwood, and Larry S. Davis. "W 4 S: A real-time system for detecting and tracking people in 2 1/2D." Computer Vision—ECCV'98. Springer Berlin Heidelberg, 1998. 877-892..
[33] Ng, Andrew Y., and Michael. I. Jordan, "On discriminative vs. generative classifiers: A comparison of logistic regression and naive Bayes." in Neural Information Processing Systems, 2001, pp.841-848.
[34] Hailang Pan, Hongwen Huo, Guoqin Cui, and Shengyong Chen, "Modeling for Deformable Body and Motion Analysis: A Review," Mathematical Problems in Engineering, vol. 2013, Article ID 786749, 14 pages, 2013. doi:10.1155/2013/786749
[35] Chaquet, J.M., E.J. Carmona, A. Fernández-Caballero, A Survey of Video Datasets for Human Action and Activity Recognition, *Computer Vision and Image Understanding* (2013), doi: http://dx.doi.org/10.1016/j.cviu.2013.01.013
[36] List, Thor, José Bins, Jose Vazquez, and Robert B. Fisher. "Performance evaluating the evaluator." In *Visual Surveillance and Performance Evaluation of Tracking and Surveillance, 2005. 2nd Joint IEEE International Workshop on*, pp. 129-136. IEEE, 2005.
[37] Guerra-Filho, Gutemberg, and Arnab Biswas. "The human motion database: A cognitive and parametric sampling of human motion." *Image and Vision Computing* 30.3 (2012): 251-261.
[38] Khurram Soomro, Amir Roshan Zamir and Mubarak Shah, UCF101: A Dataset of 101 Human Action Classes From Videos in The Wild., CRCV-TR-12-01, November, 2012.
[39] Kuehne, H., H. Jhuang, E. Garrote, T. Poggio, and T. Serre. HMDB: A Large Video Database for Human Motion Recognition. ICCV, 2011.
[40] Gkalelis, N., Hansung Kim, Adrian Hilton, Nikos Nikolaidis, and Ioannis Pitas. "The i3dpost multi-view and 3d human action/interaction database." In *Visual Media Production, 2009. CVMP'09. Conference for*, pp. 159-168. IEEE, 2009.
[41] Strat, Thomas M. "Employing contextual information in computer vision." In Proceedings of ARPA Image Understanding Workshop, 1993
[42] Aggarwal, J. K., and M. S. Ryoo. "Toward a unified framework of motion understanding." *Image and Vision Computing* 30.8 (2012): 465-466.
[43] Tsotsos, John K. "Motion understanding: Task-directed attention and representations that link perception with action." *International Journal of Computer Vision* 45.3 (2001): 265-280.
[44] Reddy, Kishore K., and Mubarak Shah. "Recognizing 50 human action categories of web videos." *Machine Vision and Applications* (2012): 1-11.
[45] Shah, Mubarak. "Guest introduction: the changing shape of computer vision in the twenty-first century." *International Journal of Computer Vision* 50.2 (2002): 103-110.
[46] Shah, Mubarak. "Understanding human behavior from motion imagery." *Machine Vision and Applications* 14.4 (2003): 210-214.
[47] Nagel, H-H. "From image sequences towards conceptual descriptions." *Image and vision computing* 6.2 (1988): 59-74.
[48] Bobick, Aaron F. "Movement, activity and action: the role of knowledge in the perception of motion." *Philosophical Transactions of the Royal Society of London. Series B: Biological Sciences* 352.1358 (1997): 1257-1265.
[49] Kambhamettu, C., D. B. Goldgof, D. Terzopoulos, and T. S. Huang, Nonrigid motion analysis, *Handbook of PRIP: Computer Vision*, Vol. 2, 1994.
[50] Chellappa, Rama, and Amit K. Roy Chowdhury. "Computer Vision, Statistics in." *Encyclopedia of Statistical Sciences*.(2006).
[51] Davis, Larry, Sandor Fejes, David Harwood, Yaser Yacoob, Ismail Hariatoglu, and Michael J. Black. "Visual surveillance of human activity." In *Computer Vision—ACCV'98*, pp. 267-274. Springer Berlin Heidelberg, 1997.
[52] Bobick, Aaron F., and James W. Davis. "The recognition of human movement using temporal templates." *Pattern Analysis and Machine Intelligence, IEEE Transactions on* 23.3 (2001): 257-267.
[53] Polana, Ramprasad, and Randal Nelson. "Low level recognition of human motion (or how to get your man without finding his body parts)." Motion of Non-Rigid and Articulated Objects, 1994., Proceedings of the 1994 IEEE Workshop on. IEEE, 1994.
[54] Ayers, Douglas, and Mubarak Shah. "Recognizing human actions in a static room." Applications of Computer Vision, 1998. WACV'98. Proceedings., Fourth IEEE Workshop on. IEEE, 1998.
[55] Oliver, Nuria M., Barbara Rosario, and Alex P. Pentland. "A Bayesian computer vision system for modeling human interactions." Pattern Analysis and Machine Intelligence, IEEE Transactions on 22.8 (2000): 831-843.
[56] Polana, Ramprasad, and Randal Nelson. "Detecting activities." Computer Vision and Pattern Recognition, 1993. Proceedings CVPR'93., 1993 IEEE Computer Society Conference on. IEEE, 1993.
[57] Weinland, Daniel, Remi Ronfard, and Edmond Boyer. "Free viewpoint action recognition using motion history volumes." Computer Vision and Image Understanding 104.2 (2006): 249-257.
[58] Yilmaz, Alper, and Mubarak Shah. "Actions As Objects: A Novel Action Representation." Computer Vision and Pattern Recognition, 2005. CVPR 2005. IEEE Computer Society Conference on. Vol. 1. IEEE, 2005.
[59] Rao, Cen, Alper Yilmaz, and Mubarak Shah. "View-invariant representation and recognition of actions." International Journal of Computer Vision 50.2 (2002): 203-226.
[60] Gorelick, Lena, et al. "Actions as space-time shapes." Pattern Analysis and Machine Intelligence, IEEE Transactions on 29.12 (2007): 2247-2253.

[61] Parameswaran, Vasu, and Rama Chellappa. "View invariance for human action recognition." International Journal of Computer Vision 66.1 (2006): 83-101.
[62] Badler, Norman I., and Stephen W. Smoliar. "Digital representations of human movement." *ACM Computing Surveys (CSUR)* 11.1 (1979): 19-38.
[63] Hampapur, Arun, Lisa Brown, Jonathan Connell, Sharat Pankanti, Andrew Senior, and Yingli Tian. "Smart surveillance: applications, technologies and implications." In *Information, Communications and Signal Processing, 2003 and Fourth Pacific Rim Conference on Multimedia. Proceedings of the 2003 Joint Conference of the Fourth International Conference on*, vol. 2, pp. 1133-1138. IEEE, 2003.
[64] Norris, Clive, Mike McCahill, and David Wood. "The Growth of CCTV: a global perspective on the international diffusion of video surveillance in publicly accessible space." *Surveillance & Society* 2.2/3 (2004).
[65] Wiliem, Arnold, Vamsi Madasu, Wageeh Boles, and Prasad Yarlagadda. "A suspicious behaviour detection using a context space model for smart surveillance systems." Computer Vision and Image Understanding 116.2 (2012): 194-209.
[66] Zhong, Hua, Jianbo Shi, and Mirkó Visontai. "Detecting unusual activity in video." Computer Vision and Pattern Recognition, 2004. CVPR 2004. Proceedings of the 2004 IEEE Computer Society Conference on. Vol. 2. IEEE, 2004.
[67] Collins, Robert, Alan Lipton, Takeo Kanade, Hironobu Fujiyoshi, David Duggins, Yanghai Tsin, David Tolliver, Nobuyoshi Enomoto, Osamu Hasegawa, Peter Burt, and Lambert Wixson. "A system for video surveillance and monitoring", Carnegie Mellon Univ., Pittsburgh, PA, Tech. Rep., CMU-RI-TR-00-12, 2000.
[68] Eng, H-L., K-A. Toh, Alvin Harvey Kam, Junxian Wang, and W-Y. Yau. "An automatic drowning detection surveillance system for challenging outdoor pool environments." In Computer Vision, 2003. Proceedings. Ninth IEEE International Conference on, pp. 532-539. IEEE, 2003.
[69] Zhan, Beibei, Dorothy N. Monekosso, Paolo Remagnino, Sergio A. Velastin, and Li-Qun Xu. "Crowd analysis: a survey." Machine Vision and Applications 19, no. 5-6 (2008): 345-357.
[70] Jacques Junior, Julio Cezar Silveira, Soraia Raupp Musse, and Claudio Rosito Jung. "Crowd analysis using computer vision techniques." Signal Processing Magazine, IEEE 27.5 (2010): 66-77.
[71] Georis, B., M. Maziere, F. Bremond, and M. Thonnat, "A video interpretation platform applied to bank agency monitoring," in Proc. 2nd Workshop Intell. Distributed Surveillance Syst., 2004, pp. 46–50.
[72] Siebel, Nils T., and S. Maybank. "The advisor visual surveillance system." ECCV 2004 workshop Applications of Computer Vision (ACV). Vol. 1. 2004.
[73] Drosou, Anastasios, et al. "Spatiotemporal analysis of human activities for biometric authentication." Computer Vision and Image Understanding 116.3 (2012): 411-421.
[74] Gafurov, Davrondzhon. "A survey of biometric gait recognition: Approaches, security and challenges." Annual Norwegian Computer Science Conference. 2007.
[75] Morimoto, Carlos H., and Marcio RM Mimica. "Eye gaze tracking techniques for interactive applications." Computer Vision and Image Understanding 98.1 (2005): 4-24.
[76] Turk, Matthew, and George Robertson. "Perceptual user interfaces." Communications of the ACM 43.3 (2000).
[77] Mikic, Ivana, Kohsia Huang, and Mohan Trivedi. "Activity monitoring and summarization for an intelligent meeting room." Human Motion, 2000. Proceedings. Workshop on. IEEE, 2000.
[78] Shotton, Jamie, Toby Sharp, Alex Kipman, Andrew Fitzgibbon, Mark Finocchio, Andrew Blake, Mat Cook, and Richard Moore. "Real-time human pose recognition in parts from single depth images." Communications of the ACM 56.1 (2013): 116-124.
[79] Zhang, Zhengyou. "Microsoft kinect sensor and its effect." Multimedia, IEEE 19.2 (2012): 4-10.
[80] Tamura, Hideyuki, Takashi Matsuyama, Naokazu Yokoya, Ryosuke Ichikari, Shohei Nobuhara, and Tomokazu Sato. "Computer vision technology applied to MR-based pre-visualization in filmmaking." In Computer Vision–ACCV 2010 Workshops, pp. 1-10. Springer Berlin Heidelberg, 2011.
[81] Assfalg, Jürgen, et al. "Semantic annotation of soccer videos: automatic highlights identification." Computer Vision and Image Understanding 92.2 (2003): 285-305.
[82] Jones, Simon, and Ling Shao. "Content-based retrieval of human actions from realistic video databases." Information Sciences (2013).
[83] Mündermann, Lars, Stefano Corazza, Ajit M. Chaudhari, Thomas P. Andriacchi, Aravind Sundaresan, and Rama Chellappa. "Measuring human movement for biomechanical applications using markerless motion capture." In Electronic Imaging 2006, pp. 60560R-60560R. International Society for Optics and Photonics, 2006.
[84] D'Orazio, Tiziana, and Marco Leo. "A review of vision-based systems for soccer video analysis." Pattern recognition 43.8 (2010): 2911-2926.
[85] Bandera, J. P., J. A. Rodríguez, L. Molina-Tanco, and A. Bandera. "A survey of vision-based architectures for robot learning by imitation." International Journal of Humanoid Robotics 9.01 (2012).
[86] Rougier, Caroline, Jean Meunier, Alain St-Arnaud, and Jacqueline Rousseau. "Robust video surveillance for fall detection based on human shape deformation." Circuits and Systems for Video Technology, IEEE Transactions on 21.5 (2011): 611-622.
[87] Chaaraoui, Alexandros André, Pau Climent-Pérez, and Francisco Flórez-Revuelta. "A review on vision techniques applied to human behaviour analysis for ambient-assisted living." Expert Systems with Applications 39.12 (2012): 10873-10888.
[88] Murphy-Chutorian, Erik, and Mohan M. Trivedi. "Head pose estimation and augmented reality tracking: An integrated system and evaluation for monitoring driver awareness." Intelligent Transportation Systems, IEEE Transactions on 11.2 (2010): 300-311.
[89] Cheng, Shinko Y., and Mohan M. Trivedi. "Turn-intent analysis using body pose for intelligent driver assistance." Pervasive Computing, IEEE 5.4 (2006): 28-37.
[90] Enzweiler, Markus, and Dariu M. Gavrila. "Monocular pedestrian detection: Survey and experiments." Pattern Analysis and Machine Intelligence, IEEE Transactions on 31.12 (2009): 2179-2195.
[91] Weng, Juyang, James McClelland, Alex Pentland, Olaf Sporns, Ida Stockman, Mriganka Sur, and Esther Thelen. "Autonomous mental development by robots and animals." *Science* 291.5504 (2001): 599-600.
[92] Yu, Chen, Linda B. Smith, Hongwei Shen, Alfredo F. Pereira, and Thomas Smith. "Active information selection: Visual attention through the hands." Autonomous Mental Development, IEEE Transactions on 1, no. 2 (2009): 141-151.
[93] Pearson, Donald E. "Developments in model-based video coding." Proceedings of the IEEE 83.6 (1995): 892-906.
[94] Akdemir, Umut, Pavan Turaga, and Rama Chellappa. "An ontology based approach for activity recognition from video." *Proceedings of the 16th ACM international conference on Multimedia*. ACM, 2008.
[95] Gruber, Tom. "Ontology." *Encyclopedia of Database Systems*, Springer-Verlag, 2009.
[96] Gruber, Thomas R. "Toward principles for the design of ontologies used for knowledge sharing." *International journal of human computer studies* 43.5 (1995): 907-928.
[97] Smith, Barry. "Ontology." *The Blackwell guide to the philosophy of computing and information* (2003): 155-166.
[98] Martin, David, Charless Fowlkes, Doron Tal, and Jitendra Malik. "A database of human segmented natural images and its application to evaluating segmentation algorithms and measuring ecological statistics." In *Computer Vision, 2001. ICCV 2001. Proceedings. Eighth IEEE International Conference on*, vol. 2, pp. 416-423. IEEE, 2001.
[99] Chen, Cheng, and Guoliang Fan. "What can we learn from biological vision studies for human motion segmentation?." *Advances in Visual Computing*. Springer Berlin Heidelberg, 2006. 790-801.

[100] Horvath, Gabor, Etelka Farkas, Ildiko Boncz, Miklos Blaho, and Gyorgy Kriska. "Cavemen Were Better at Depicting Quadruped Walking than Modern Artists: Erroneous Walking Illustrations in the Fine Arts from Prehistory to Today." *PLoS one* 7.12 (2012): e49786.
[101] Maringer, Johannes. "Adorants in Prehistoric Art: Prehistoric Attitudes and Gestures of Prayer." *Numen* 26.2 (1979): 215-230. Stable URL: http://www.jstor.org/stable/3269720
[102] Klette, Reinhard, and Garry Tee. *Understanding human motion: A historic review*. Springer Netherlands, 2008.
[103] Pipes, Alan. "*Foundations of art+ design.*" Laurence King Publishing, 2003.
[104] Cutting, James E. "Representing motion in a static image: constraints and parallels in art, science, and popular culture." *PERCEPTION-LONDON-* 31.10 (2002): 1165-1194.
[105] Joshi, Alark, and Penny Rheingans. "Illustration-inspired techniques for visualizing time-varying data." *Visualization, 2005. VIS 05. IEEE.* IEEE, 2005.
[106] Cavanagh, Patrick. "The artist as neuroscientist." *Nature* 434.7031 (2005): 301-307.
[107] Watts, Edith, and Barry Girsh. *Art of Ancient Egypt: A Resource for Educators*. Metropolitan Museum of Art, 1998.
[108] "Egyptian religion". *Encyclopædia Britannica. Encyclopædia Britannica Online.* Encyclopædia Britannica Inc., 2013. Web. 12 Apr. 2013 <http://www.britannica.com/EBchecked/topic/180764/Egyptian-religion>.
[109] Soper, Alexander C. "Life-motion and the Sense of Space in Early Chinese Representational Art." *The Art Bulletin* 30.3 (1948): 167-186. Stable URL: http://www.jstor.org/stable/3047182 .
[110] Inal, Güner. "artistic relationship between the far and the near east as reflected in the miniatures of the ǧāmi at-tawārīḫ." *Kunst des Orients* 10.1/2 (1975): 108-143.
[111] Shan, Gongbing. "Signatures of Human Movements–Understand Gestural Content Using Motion Capture Trajectories." Journal of Creative Work, Volume 2, Issue 1, 2008.
[112] Aristotle, Jonathan Barnes. "Complete works of Aristotle." *Ed. J. Barnes, Princeton, NJ* (1995).
[113] Arterberry, M. E. "Perceptual development. " In M. M. Haith & J. B. Benson (Eds.), *Encyclopedia of Infant and Early Childhood Development*. San Diego: Elsevier. (2008).
[114] Blake, Randolph, and Maggie Shiffrar. "Perception of human motion." *Annu. Rev. Psychol.* 58 (2007): 47-73.
[115] Kubovy, M., S. E. Palmer, M. A. Peterson, M. Singh, and R. von der Heydt. "A Century of Gestalt Psychology in Visual Perception." *Psychological Bulletin* , Vol. 138. 6, (2012):1172–1217.
[116] Rock, Irvin, and Stephen Palmer. "The legacy of Gestalt psychology." *Scientific American* 263.6 (1990): 84-90.
[117] Peterson, Mary, and Elizabeth Salvagio. "Figure-ground perception." *Scholarpedia* 5.4 (2010): 4320. doi:10.4249/scholarpedia.4320
[118] Koffka, Kurt. "Principles of Gestalt psychology." (1935): 221.
[119] "Gestalt psychology". *Encyclopædia Britannica. Encyclopædia Britannica Online.* Encyclopædia Britannica Inc., 2013. Web. 10 Apr. 2013 <http://www.britannica.com/EBchecked/topic/232098/Gestalt-psychology>.
[120] Cutting, James E. "Four assumptions about invariance in perception." *Journal of Experimental Psychology: Human Perception and Performance* 9.2 (1983): 310-317.
[121] Johansson, Gunnar. "Visual perception of biological motion and a model for its analysis." *Perception & Psychophysics* 14.2 (1973): 201-211.
[122] Kozlowski, Lynn T., and James E. Cutting. "Recognizing the sex of a walker from a dynamic point-light display." *Perception & Psychophysics* 21.6 (1977): 575-580.
[123] Cutting, James E., and Lynn T. Kozlowski. "Recognizing friends by their walk: Gait perception without familiarity cues." *Bulletin of the psychonomic society* 9.5 (1977): 353-356.
[124] Pinker, Steven. "Visual cognition: An introduction." *Cognition* 18.1 (1984): 1-63.
[125] Tomasello, Michael, Malinda Carpenter, Josep Call, Tanya Behne, and Henrike Moll. "Understanding and sharing intentions: The origins of cultural cognition." *Behavioral and brain sciences* 28.5 (2005): 675-690.
[126] Rizzolatti, Giacomo, Leonardo Fogassi, and Vittorio Gallese. "Neurophysiological mechanisms underlying the understanding and imitation of action." *Nature Reviews Neuroscience* 2.9 (2001): 661-670.
[127] Giese, Martin A., and Tomaso Poggio. "Neural mechanisms for the recognition of biological movements." *Nature Reviews Neuroscience* 4.3 (2003): 179-192.
[128] Rizzolatti, Giacomo, and Laila Craighero. "The mirror-neuron system." *Annu. Rev. Neurosci.* 27 (2004): 169-192.
[129] Rizzolatti, Giacomo, and Maddalena Fabbri-Destro. "The mirror system and its role in social cognition." *Current opinion in neurobiology* 18.2 (2008): 179.
[130] Mukamel, Roy, Arne D. Ekstrom, Jonas Kaplan, Marco Iacoboni, and Itzhak Fried. "Single-neuron responses in humans during execution and observation of actions." *Current biology* 20. 8 (2010): 750-756.
[131] Marčelja, S. "Mathematical description of the responses of simple cortical cells*." *JOSA* 70.11 (1980): 1297-1300.
[132] Gabor, Dennis. "Theory of communication. Part 1: The analysis of information." *Electrical Engineers-Part III: Radio and Communication Engineering, Journal of the Institution of* 93.26 (1946): 429-441.
[133] Daugman, John G. "Uncertainty relation for resolution in space, spatial frequency, and orientation optimized by two-dimensional visual cortical filters." *Optical Society of America, Journal, A: Optics and Image Science* 2.7 (1985): 1160-1169.
[134] Jones, Judson P., and Larry A. Palmer. "An evaluation of the two-dimensional Gabor filter model of simple receptive fields in cat striate cortex." *Journal of Neurophysiology* 58.6 (1987): 1233-1258.
[135] Marr, David. "Vision: A computational investigation into the human representation and processing of visual information, Henry Holt and Co." *Inc., New York, NY* (1982).
[136] Marr, David, and Herbert Keith Nishihara. "Representation and recognition of the spatial organization of three-dimensional shapes." *Proceedings of the Royal Society of London. Series B. Biological Sciences* 200.1140 (1978): 269-294.
[137] Marr, David, and Lucia Vaina. "Representation and recognition of the movements of shapes." *Proceedings of the Royal Society of London. Series B. Biological Sciences* 214.1197 (1982): 501-524.
[138] Glennerster, Andrew. "Marr's vision: Twenty-five years on." *Current biology* 17.11 (2007): R397-R399.
[139] Edelman, Shimon, and Lucia M. Vaina. "David marr." *International encyclopedia of the social and behavioral sciences* (2001).
[140] Haggard, Patrick. "Conscious intention and motor cognition." *Trends in cognitive sciences* 9.6 (2005): 290-295.
[141] Smith, Kerri. "Taking aim at free will." *Nature* 477.7362 (2011): 23-25.
[142] Soon, Chun Siong, Marcel Brass, Hans-Jochen Heinze, and John-Dylan Haynes. "Unconscious determinants of free decisions in the human brain." *Nature neuroscience* 11. 5 (2008): 543-545.
[143] http://www.nada.kth.se/cvap/actions/ (Last accessed March 2014)
[144] http://www.di.ens.fr/~laptev/actions/hollywood2/ (Last accessed March 2014)
[145] http://vision.stanford.edu/Datasets/OlympicSports/ (Last accessed March 2014)
[146] http://serre-lab.clps.brown.edu/resource/hmdb-a-large-human-motion-database/ (Last accessed March 2014)
[147] http://crcv.ucf.edu/data/UCF101.php (Last accessed March 2014)
[148] http://www.robots.ox.ac.uk/~vgg/data/tv_human_interactions/ (Last accessed March 2014)
[149] http://kahlan.eps.surrey.ac.uk/i3dpost_action/ (Visited May 2013)

[150] http://www.cbsr.ia.ac.cn/english/Action%20Databases%20EN.asp (Last accessed March 2014)
[151] http://www.cvg.rdg.ac.uk/datasets/index.html (Last accessed March 2014)
[152] http://www.cvg.rdg.ac.uk/PETS2013/a.html (Last accessed March 2014)
[153] http://all-history.org/2-3.html (Visited March 2013)
[154] Rick Doble. "Historic Timeline: The Capture of Movement in Painting and Photography - Illustrated sequence from da Vinci to today", September 7, 2011 http://www.pixiq.com/article/historic-timeline-capture-of-movement-in-painting-photography (Visited March 2013)
[155] http://www.drawingsofleonardo.org/ (Last accessed March 2014)
[156] http://www.qotd.org/search/search.html?author=Michelangelo&keywords=&page=2 (Last accessed March 2014)
  OR http://quote.robertgenn.com/auth_search.php?name=Michelangelo (Last accessed March 2014)
[157] http://www.qotd.org/search/search.html?author=edgar%20degas&keywords=&page=3 (Last accessed March 2014)
  OR http://quote.robertgenn.com/auth_search.php?name=Edgar+Degas (Last accessed March 2014)